\documentclass[journal]{IEEEtran}
\usepackage[dvipsnames, svgnames, x11names]{xcolor}
\usepackage{graphicx}
\usepackage{amsmath}
\usepackage{amssymb}
\usepackage{multirow}
\usepackage{color}
\ifCLASSINFOpdf
\else
\fi
\begin{document}
	
	\title{Incomplete Descriptor Mining  with Elastic Loss for  Person Re-Identification}

	\author{Hongchen Tan,  Xiuping Liu*,  Yuhao  Bian,    Huasheng  Wang,    and   Baocai Yin

		
		\thanks{(Corresponding author: Xiuping Liu)
			
			Hongchen Tan, Xiuping Liu,  Yuhao Bian, and  Huasheng Wang    are with School of Mathematical Sciences, Key Laboratory of Computational Mathematics,  Dalian University of Technology, Dalian 116024, China (e-mail: tanhongchenphd@mail.dlut.edu.cn/tanhongchenphd@126.com; xpliu@dlut.edu.cn; yhbian@mail.dlut.edu.cn; huashengdadi@mail.dlut.edu.cn.).

			Baocai Yin is with the Department of Electronic Information and Electrical Engineering,   Dalian University of Technology, Dalian 116024, China (e-mail: ybc@dlut.edu.cn).
			
		\textbf{Copyright@2021 IEEE. Personal use of this material is permitted. However, permission to use this material for any other purposes must be obtained from the IEEE by sending an email to pubs-permissions@ieee.org.}	
				
		}
		\thanks{}
		\thanks{}}

	\markboth{Journal of \LaTeX\ Class Files,~Vol.~14, No.~8, August~2015}%
	{Shell \MakeLowercase{\textit{et al.}}: Bare Demo of IEEEtran.cls for IEEE Journals}

	\maketitle
\begin{abstract}
In this paper, we propose a novel person Re-ID model,  Consecutive  Batch DropBlock Network (CBDB-Net),   to capture  the attentive  and  robust person descriptor for  the person Re-ID  task. The CBDB-Net  contains  two novel  designs:  the Consecutive  Batch DropBlock  Module (CBDBM)  and  the Elastic Loss (EL). 
In the  Consecutive  Batch DropBlock  Module (CBDBM),  we  firstly   conduct  uniform  partition on  the feature maps. And then,  we  independently  and  continuously  drop each  patch  from top to bottom  on the  feature maps, which can output  multiple incomplete  feature maps. 
In  the training stage, these  multiple incomplete  features  can better encourage  the Re-ID model   to capture  the robust  person descriptor  for  the Re-ID task.  
In  the  Elastic Loss (EL),   we  design a  novel weight control item to   help  the Re-ID model adaptively  balance  hard  sample  pairs  and easy sample pairs  in the  whole  training  process.  
Through an extensive set of ablation studies, we verify that the Consecutive  Batch DropBlock Module (CBDBM) and  the Elastic Loss (EL)   each contribute to the performance boosts of CBDB-Net.   
We  demonstrate that   our CBDB-Net  can achieve   the competitive   performance  on the three standard person Re-ID datasets (the  Market-1501, the DukeMTMC-Re-ID,  and the  CUHK03  dataset),  three occluded  Person Re-ID  datasets (the Occluded DukeMTMC, the Partial-REID,  and  the Partial iLIDS dataset), and  a  general  image  retrieval dataset (In-Shop Clothes Retrieval dataset).

\end{abstract}


\IEEEpeerreviewmaketitle
\section{Introduction}~\label{Introduction}


Person Re-Identification (Re-ID) has attracted increasing attention from both the academia and the industry  due to its significant role in the video surveillance. 
Given  a target person  image captured by one camera, the goal of  the  person Re-ID task  is to re-identify the same  person  from   images captured by
other  cameras'   viewpoints. 
Despite the  exciting  progress in recent years, the person Re-ID task remains to be extremely challenging.
This is  because the  task is  easily affected by  body misalignment, occlusion, background perturbance, and viewpoint changes, etc.  
To tackle this  challenge,   almost person Re-ID approaches focus  on  two  strategies:  feature  descriptor learning and distance metric learning. 
The  former  approaches aim  to capture the discriminative  person   descriptor  which  is robust  to various interference factors; The latter approaches  aim to gain a  better  metric space  equipped with  better  classification discrimination of different persons.

Recently, many  outstanding approaches~\cite{Zuozhuo19, Geng2016Deep,  Coherence_Hierarchical2019Person, HiGCIN_Hierarchical2019Person, Hierarchical2019Person, Lin2017Improving, Zheng2016Person, Varior2016Gated, Cheng2016Person, 2018Sharp, 2017Pedestrian, 2017Deep}, equipped  with  a series  of  metric constraints,  try  to  capture the global descriptor of  the whole person image.   However, the   global  descriptors are  easily prone  to the false matching between persons who look similar,   due  to  lacking    enough local  information.  
Therefore,  other  approaches~\cite{Yeong-Jun16, Kalayeh2018Human, Chunfeng, Xuelin18, Chi17, Spatiotemporall2019Person, Jing181}  introduce the pose estimation  or human parsing  models   to  help   the person Re-ID  model  better locate  and  capture  local human part feature.   
However, the underlying datasets bias between pose estimation, human parsing,  and person  Re-ID  remains an obstacle against the ideal human body  partition on person images.  
And the  additional  human  models   make  the  person Re-ID model  more  complex  and unwieldy.  
Thus, in  this  paper,  we  aim   to design a  simple and  effective  strategy  to  help  the  person Re-ID model capture a  high-quality  person descriptor.

\begin{figure}\centering
	\begin{center} 
		\includegraphics[scale=0.5]{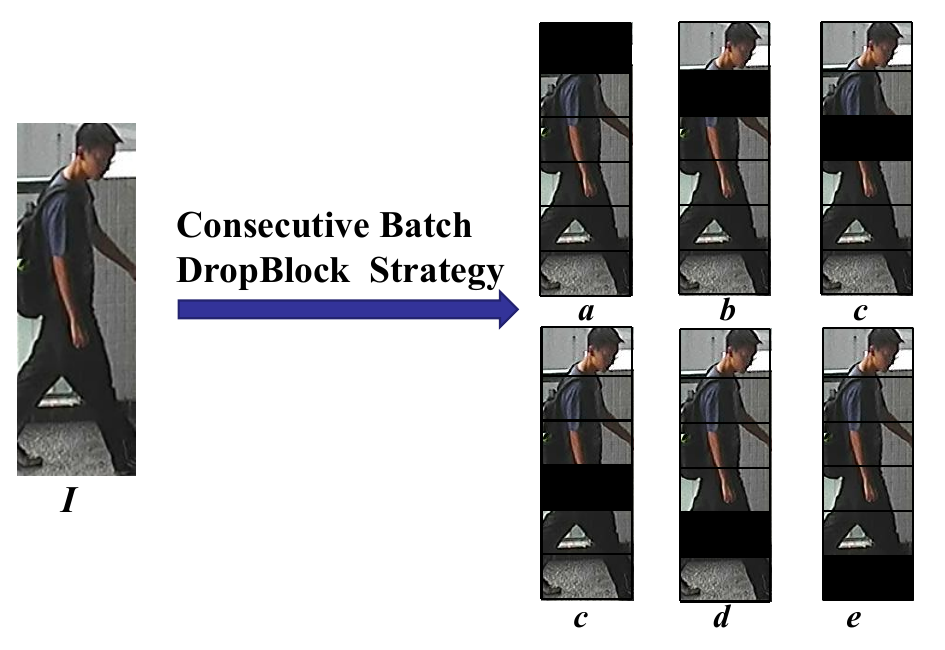}
	\end{center} 
	\caption{To easily understand   the   Consecutive Batch DropBlock (CBDB) strategy,  we  directly use the raw  person image to show  the operation  of our  CBDB strategy.  In  fact, the Consecutive Batch DropBlock (CBDB) strategy  is  conducted  on the conv-layer.} 
	\label{tcsvt-1} 
\end{figure}

Recently, the Dropout strategies~\cite{Jonathan2015, Nitish2014, Golnaz2018, Zuozhuo19, ZhunZhong2017, Terrance2017},  is widely used in   person Re-ID  and  other visual tasks.  Compared  with  the additional human model assistance,  these Dropout strategies are  the lightweight  and  effective  for person  Re-ID models.  It can better help the  person Re-ID model embed in the application device.   In these Dropout strategies, Cutout~\cite{Terrance2017}, random erasing~\cite{ZhunZhong2017},  and SpatialDropout~\cite{Jonathan2015},  can randomly drop  the feature pixel in  the feature maps  or feature  vectors. 
However, these methods  only  belong to  a  regularization method  and not attentive feature learning methods. 
And  they  can  not drop   a large contiguous area within  a training batch. 
So,  Batch DropBlock~\cite{Zuozhuo19}  is  proposed   to drop  the same continuity region in a training batch.     Inspired by  the Batch DropBlock strategy in~\cite{Zuozhuo19},   our first idea is  to propose a novel  drop strategy,  Consecutive Batch DropBlock (CBDB), to produce  multiple incomplete  feature maps for  capturing  high-quality  person descriptor.

Different  from  Batch DropBlock: \textbf{(i)} our  Consecutive Batch DropBlock strategy can gain  multiple  incomplete  feature  maps to train the deep  Re-ID model; \textbf{(ii)} our consecutive Batch DropBlock drops  the same patch for the whole training set instead of a  batch.  
As  shown in the Figure~\ref{tcsvt-1}, for  the Consecutive Batch DropBlock,  we  firstly  conduct  uniform partition on  the conv-layer.    Secondly, we  independently  and  continuously  drop each  patch  from top to bottom  on the  conv-layer.   Since  this,   we  can gain multiple  incomplete   feature maps  to  push   the deep  model to capture  the  robust  and high-quality person  feature  descriptor.   Thus,  our Consecutive Batch DropBlock  is  an  effective attentive feature learning strategy. Besides,  the Consecutive Batch DropBlock strategy can be regarded as  the Batch DropBlock'~\cite{Zuozhuo19}  extension, and  the  complementary  strategy  to uniform patch descriptors in PCB~\cite{Yifan2018}.   
Thus, we  believe  that the   Consecutive Batch DropBlock (CBDB)  strategy  can effectively improve the robustness  of  person  descriptors  in  the  person Re-ID task.

Based  on the Consecutive Batch DropBlock, we  can gain   multiple  incomplete feature maps (descriptors)   in the training  process. 
However,  inevitably there will be hard sample  pairs  and easy sample pairs  for  the  person matching task.  
The Batch Hard Triplet  Loss~\cite{Alexander2017}  may be a suitable  metric loss  function to  balance  these  hard sample pairs. 
However, in the whole  training  process, the  difficulty level  of hard sample pairs  are  different  in different training stage; In the whole  training sets, the  difficulty level  of hard sample pairs  are  also  different  in different  person ID.       
So, our  second  idea  is  to design  a  novel  metric  loss  function to dynamically  balance  the  hard sample pairs  and  easy sample pairs    in  the training process.

Based on the  above  analysis, we  propose a novel  person Re-ID model,  Consecutive  Batch DropBlock Network (CBDB-Net). The CBDB-Net  contains  two  novel designs: the  Consecutive  Batch DropBlock  strategy  and  the  Elastic Loss.  The  former  exploits  multiple  incomplete  descriptors  to  improve  the robustness of  the deep model.  And, in  the  testing stage, a simple test model is adopted  to produce  a  high-quality  person descriptor for  the person matching.  The latter  can  better    mine and balance  the hard  sample  pairs  for  the  whole training samples  in the  whole  training  process.  It  can  further improve  the performance  of  the deep person Re-ID  model.

In the experimental section, firstly we  validate our CBDB-Net on three  generic  person Re-ID datasets: Market-1501~\cite{Zheng2015Scalable}, DukeMTMC-reID~\cite{Ristani2016Performance, ZhedongZheng}, CUHK03~\cite{Weireid2014}.	
Secondly,  we evaluate  our CBDB-Net on three occluded  person Re-ID datasets: Occluded-DukeMTMC~\cite{Jiaxu112019},  Partial-REID~\cite{WeiShi2015}, and  Partial-iLIDS~\cite{Lingxiao2018}. 
Finally,  we believe our CBDB-Net can be applied to other  image retrieval  tasks. So, we  evaluate our CBDB-Net  on the In-Shop Clothes retrieval dataset~\cite{Ziwei2016}.
Extensive experimental results and analysis demonstrate the effectiveness of CBDB-Net and significantly improved performance compared against  most 
state of the arts over two evaluation metrics.

\begin{figure*}\centering
	\begin{center} 
		\includegraphics[scale=0.35]{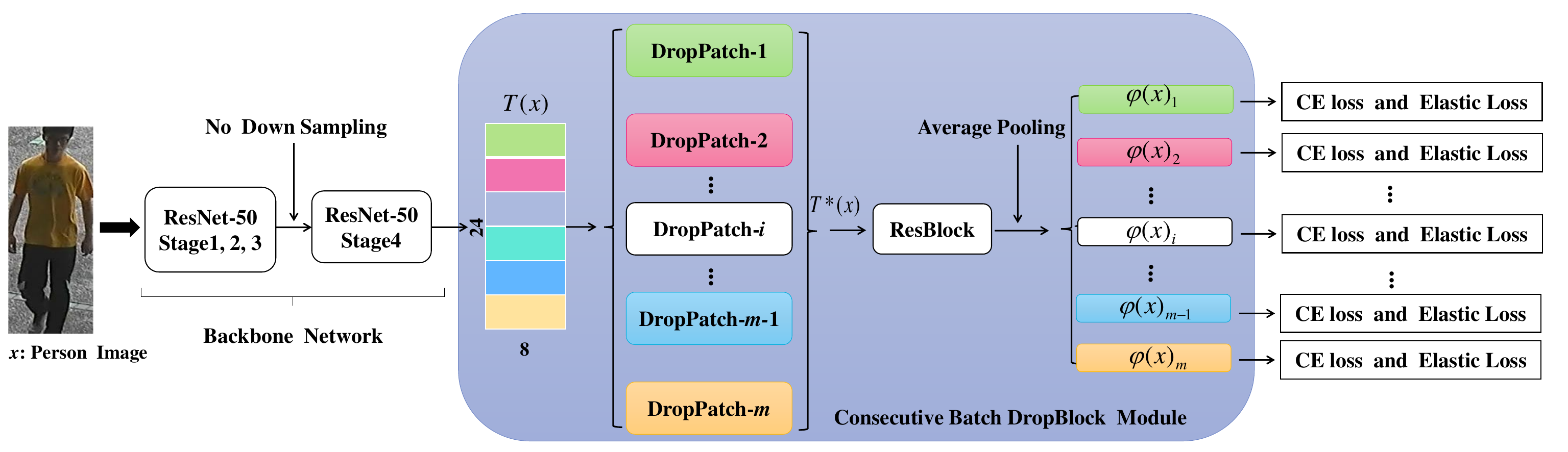}
	\end{center} 
	\caption{The architecture of Consecutive  Batch DropBlock Network (CBDB-Net) for the person Re-ID task.   The  two  novel strategies are Consecutive  Batch DropBlock Module and  the  proposed Elastic Loss.  Here  the ``CE'' denotes the  Cross-Entropy  Loss function.    } 
	\label{tcsvt-2} 
\end{figure*}

\begin{figure}\centering
	\begin{center} 
		\includegraphics[scale=0.32]{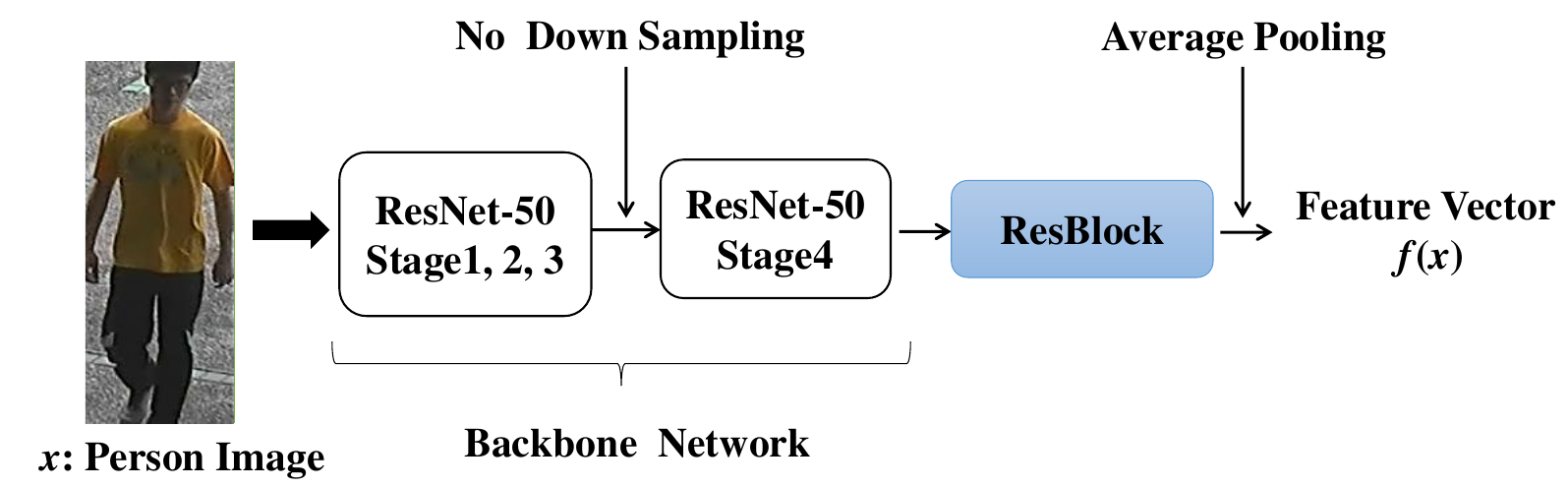}
	\end{center} 
	\caption{The network  pipeline of the proposed CBDB-Net   in the testing stage. } 
	\label{tcsvt-5} 
\end{figure}

\section{Related  Work}

\subsection{Part-based  person Re-ID Models}

In our CBDB-Net,  the Consecutive  Batch DropBlock  strategy  can  guide    the  person  Re-ID model  to gain   multiple  incomplete feature  maps  of one  person  image  in  the  training  process. 
These   incomplete feature  maps  can be  regarded as   the large  parts.  
So, in this subsection, we introduce  related works of the part-based  person Re-ID task. 
Recently, ~\cite{Kalayeh2018Human, Saquib2018, Hierarchical2019Person, Chi17, Haiyu2017} adopted  an additional  human body part detector or an additional human body  parsing model to focus  on  more accurate human parts. E.g., SPReID~\cite{Kalayeh2018Human} applied  an additional human body  parsing model to generate $5$ different pre-defined body part masks to capture more reliable part representations.  
~\cite{Jianyuan19} addressed the missed contextual information by exploiting both the accurate human body parts and the coarse non-human parts.  ~\cite{Jiaxu112019, Guanan2020}  combined   the  pose landmarks  and  uniform  partial feature  to improve   the  performance  of  the occluded person  Re-ID  task.  
~\cite{Longhui2017}  adopted  the key  point  detector  to exploit three  coarse  body part,  and combine the  global information to  conduct   the  person matching.     
Different  from  ~\cite{Longhui2017},   ~\cite{Yanbeiarxiv2017} dropped  the keypoint detector, only  used   the maximum  feature responses  to  locate  the body  regions  and combined with  the Part  Loss. 
~\cite{Yifan2018} conducted  uniform partition on  the conv-layer for learning part-level features.  
Our CBDB-Net  also  belongs  to    part-based  person Re-ID methods to  some degree.  
Similar  to ~\cite{Yifan2018, Yanbeiarxiv2017}, we  also  needn't  any  additional human  models'  assistance.  
In  the  CBDB-Net,   we can   gain  multiple  large parts  by  the  proposed  Consecutive  Batch DropBlock  strategy. These  large  feature  parts  can be regarded  as   complementary to   the uniform  parts in~\cite{Yifan2018, Yanbeiarxiv2017}.  And, compared  with  the small feature parts  in  ~\cite{Longhui2017, Yifan2018, Yanbeiarxiv2017},  these  large feature parts can  better  encourage  the deep  Re-ID  model to capture the robust  person structual  information.

\subsection{Triplet  Ranking in person Re-ID}

Triplet Ranking Loss~\cite{Florian2015} is one of  the most  important metric loss functions,  which  encourages the distance between positive sample  pairs  to be closer than negative sample pairs.  It has been applied in various outstanding  deep  vision  models,  and  achieves  outstanding performance in these  metric learning  tasks. \cite{triplet2015}  may be the  first one to introduce the Triplet Loss into the Re-ID task. When   SPGAN~\cite{Deng2017Image}  conducted  the cross-domain person Re-ID  task,   they   adopted  the Contrastive Loss  to  preserve  the person ID information in the cross-domain image style transfer.   ~\cite{Cheng2016Person}  extended the Triplet Loss by introducing the  absolute distance of  the  positive  sample pair.   
~\cite{Sanping19}  proposed  a virtual sample  in the triplet unit to accelerate  sample distance  optimization.
Similar to ~\cite{VSEFartash2018}, ~\cite{Alexander2017}  proposed  the batch  hard  Triplet Loss  by  introducing   the hard  sample mining strategy for  person  sample  pairs.  Since  this,  many  SOTA  Re-ID  methods~\cite{Kaiyang19, Tianlong2019, Bryan2019, Wenjie19, Ruibing19, Zuozhuo19}  adopted the Batch  Hard Triplet Loss to   gain the  outstanding performance.

Inspired  by  these  methods~\cite{Kaiyang19, Tianlong2019, Bryan2019, Wenjie19, Ruibing19, Zuozhuo19},  we  also introduce the hard  sample mining strategy   to design  a novel triplet  loss in our  CBDB-Net. 
Different from the original Batch Hard Triplet Loss~\cite{Alexander2017},  our   proposed  triplet loss  can  dynamically  adjust  the learning  weights  of  hard sample  pairs  in the  whole  training  process.   
The  experiments  show  that  it   can  further   help  the  person Re-ID model   to  gain  better  performance.


\section{CBDB-Net}

In   this section, we  describe  the details of the proposed Consecutive  Batch DropBlock Network (CBDB-Net).   
The  training  framework  is  shown in Figure~\ref{tcsvt-2};  The  testing framework  is  shown in Figure~\ref{tcsvt-5}. 

\textbf{In the  training stage.} As  shown  in Figure~\ref{tcsvt-2},  our CBDB-Net  contains three  components: the Backbone Network,  the Consecutive Batch DropBlock  Module (CBDBM),  and the loss functions (which  contain  our proposed Elastic Loss).   
The  input  to  the   Backbone Network  is the  person  image  $x$. 
The Backbone Network  provides  the basic feature  maps  $T(x)$ for   the Consecutive Batch DropBlock  Module (CBDBM). 
The core  of   the   Consecutive Batch DropBlock  Module (CBDBM) is  the  Consecutive Batch DropBlock  strategy.    The input  of  the CBDBM is  the basic feature  maps  $T(x)$.  The CBDBM  outputs   multiple incomplete person  descriptors $\varphi(x)_i, i=1, 2, \cdots, m$. The  loss  functions constraint  the  whole  CBDB-Net  to capture  the  robust and  high-quality  person  information for the person matching task. 
The  loss  functions  contain  the Cross-Entropy Loss and our  proposed  Elastic Loss.

\textbf{In the  testing stage.}  As  shown  in Figure~\ref{tcsvt-5},  the  testing framework   contains the Backbone Network,  ResBlock,   and Average  Pooling Operation. 
We use  the  feature $f(x) \in \mathbb{R}^{512}$   to  find the best matching person in the gallery by comparing the squared distance, i.e. $d(a, b) =\|a-b\|^{2}_{2}$.

In the  following parts, we mainly  describe  the  training framework of  our CBDB-Net.  We  describe  the details  of Backbone Network  in Subsection~\ref{Backbone}, the details  of Consecutive Batch DropBlock  Module (CBDBM)   in Subsection~\ref{CBDBMSECTION},   the details  of   the proposed Elastic Loss   in Subsection~\ref{Elastic}, and the Network Architecture Overview~\ref{NAO}.

\subsection{Backbone Network}~\label{Backbone}

Following current many outstanding methods~\cite{Kaiyang19, Tianlong2019, Bryan2019, Wenjie19, Ruibing19, Zuozhuo19},  our  CBDB-Net also uses the ResNet-50~\cite{KaimingRSE2015}  pre-trained on ImageNet~\cite{Imagenet2009} as  the backbone network,   to encode a  person  image $x$.   To get  a larger size high-level feature tensor, we also  modify the basic structure of the ResNet-50 slightly. The down-sampling operation at the beginning of the  ``ResNet-$50$ Stage $4$'' is not employed.  Therefore, we can get a larger feature tensor  $T(x) \in \mathbb{R}^{24 \times 8 \times 2048}$.

\subsection{Consecutive  Batch DropBlock  Module}~\label{CBDBMSECTION}

 \begin{figure}\centering
	\begin{center} 
		\includegraphics[scale=0.55]{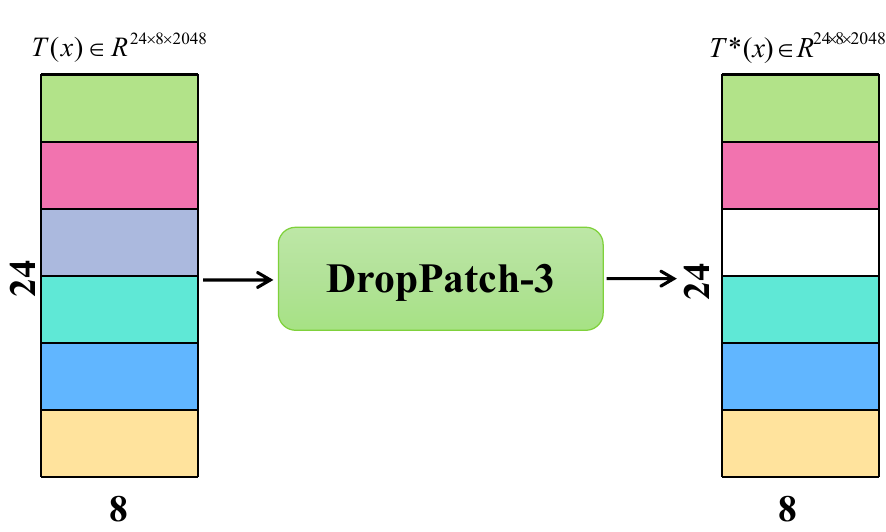}
	\end{center} 
	\caption{ An  example of $DropPatch-3$'s  operation.   The DropPatch-$3$ is  used   to drop  the $3$-th patch on  the tensor $T(x)$. So,  we  can see  that    the $3$-th  patch on  the $T(x)$ are zeroed out.} 
	\label{tcsvt-3} 
\end{figure}

Based on the large  feature tensor $T(x) \in \mathbb{R}^{24 \times 8 \times 2048}$,   we  build  the  Consecutive Batch DropBlock  Module (CBDBM). 

\textbf{(i:)} As  shown in Figure~\ref{tcsvt-2},  the   tensor $T(x)$   is  divided  to  $m$  uniform  patches.  
\textbf{(ii:)}  The $DropPatch-i, \emph{i} = 1,2,\cdots, m$ is designed  to drop  the $i-th$ patch on   the tensor $T(x)$. 
As  shown in the  Figure~\ref{tcsvt-3},     the  feature tensor $T(x)$  is  divided  to $6$  uniform  patches;   
The $DropPatch-3$ is  used   to drop  the $3-th$ patch on  the tensor $T(x)$. We  can see  that    the $3-th$ patch on  the $T(x)$ is zeroed out.   
Since  this,   we can  gain  $m$ incomplete  feature tensors  $T(x)^{*}_i \in \mathbb{R}^{24 \times 8 \times 2048},  \emph{i} = 1,2,\cdots, m$. 
\textbf{(iii:)} These  $m$ incomplete  feature tensors  $T(x)^{*}_i \in \mathbb{R}^{24 \times 8 \times 2048},  \emph{i} = 1,2,\cdots, m$  are  fed into the   ``ResBlock''  and  average  pooling  operation. Since  this,   we  can gain  $m$  incomplete  person  descriptors, i.e.  $\varphi(x)_i \in \mathbb{R}^{512},  \emph{i} = 1,2,\cdots, m$  which   is  fed  into the loss functions: the Cross-Entropy Loss  and the  proposed Elastic Loss~\ref{Elastic}.   Here,  the  ``ResBlock'' is  composed  of  three   bottleneck blocks~\cite{KaimingRSE2015}. 

\textbf{Analysis.} 
Based  on the  person  feature map    $T(x)$,  one patch  feature region  on  the $T(x)^{*}_i$ is  missed.  In  the  training stage,  the person Re-ID loss functions can   push  the model to capture key person  information for Re-ID from the remaining areas on the $T(x)^{*}_i$. Our  CBDBM  can  produce multiple  incomplete feature maps  $T(x)^{*}_i,  \emph{i} = 1,2,\cdots, m$. Since this,  the deep  Re-ID model is required to adapt to different  patch  dropout  cases  to capture key person information.  In  the testing stage, as  shown in Figure~\ref{tcsvt-5}, the CBDBM  is moved. 
Thus, the  deep  Re-ID  model can better extract robust  and key  person  information from the whole  person image feature.

\subsection{The proposed Elastic Loss}~\label{Elastic} 

The  CBDBM  ouputs  $m$  incomplete descriptors  $\varphi(x)_i \in \mathbb{R}^{512},  \emph{i} = 1,2,\cdots, m$ of  one  person  image $x$.  In the  training process, these incomplete descriptors  inevitably  contain   many  hard matching sample  pairs. 
Recently,  the Batch Hard Triple (BHT) Loss~\cite{Alexander2017}   introduced  the  hard sample mining strategy  to effectively focus on   the hard sample  pairs in the  training  process.   However,   the hard sample mining strategy  in the   Batch  Hard Triple Loss~\cite{Alexander2017}   did  not consider  the  two  issues: \textbf{(a:)}  in the  different  training stage,  the difficulty level  of hard samples  pairs are  different;   \textbf{(b:)} in  each  training step or  epoch,   the difficulty level of hard samples  pairs from  variant ID person are also different.   
Recently, Focal  Loss~\cite{Lin2017Focal} introduced  the weight control item into  the Cross-Entropy Loss, which  can  dynamically adjust  the weight of   hard samples and  easy  samples in the training process. Inspired  by  the  Focal  Loss~\cite{Lin2017Focal}  and the   Batch Hard Triple Loss~\cite{Alexander2017},  we   propose  a  novel Elastic Loss  to relieve  the above  two  issues.

To define the Elastic Loss, we \textbf{firstly} organize the training samples into a set of triplet feature units, $S = {(s(x^{a}), s(x^{p}), s(x^{n}))}$ which simply denotes as  $S = {(s^{a}, s^{p}, s^{n})}$, and   the raw  person image triplet  units  is  $X = {(x^{a}, x^{p}, x^{n})}$.
Here, $(s^{a}, s^{p})$ represents a positive pair's feature with $y^{a}=y^{p}$, and $(s^{a}, s^{n})$ indicates a negative pair's feature with $y^{a} \neq y^{n}$.   
Here, $y\in Y$ is the person ID information. 
We  use  $d(a, b) =\|a-b\|^{2}_{2}$ to denote the squared distance  between the  feature  vectors $a$  and  $b$ in the  feature space.

\textbf{Secondly},  we  revisit  the  Batch Hard Triplet Loss~\cite{Alexander2017}.  
Based on the Triplet Loss~\cite{Florian2015}, ~\cite{Alexander2017} extended  the Triplet Loss~\cite{Florian2015}   by introducing  the hard sample mining strategy. 
Here, the hard sample mining strategy in  the training batch:   the positive sample pair with the largest distance as the hard positive sample pair; the negative sample pair with the smallest distance as the hard negative sample pair. 
In  the training  process,   the hard  sample  pair  will be focused on.   
Based on the design,   the Batch Hard Triple Loss function is defined as:
\begin{equation}~\label{hardtriplet}
	\mathcal{T}_{HardTriplet}= [\eta+ \underset{x^{a},x^{p}}{\text{max}} d(s^{a}, s^{p})- \underset{x^{a},x^{n}}{\text{min}} d(s^{a}, s^{n})]_{+} 
\end{equation}
Here, $\eta$ represents the margin parameter.

\textbf{Finally}, we  define  the Elastic Loss by  revising  the  Batch Hard Triplet  Loss.  
Same  as  the  Eq.~\ref{hardtriplet},  the  hard positive sample pairs  is the maximum distance  between $s^{a}$  and $s^{p}$, i.e. $\underset{x^{a},x^{p}}{\text{max}} d(s^{a}, s^{p})$,   within the  training  batch;   and  the hard negative sample pairs  is the  minimum  distance  between $s^{a}$  and $s^{n}$, i.e. $\underset{x^{a},x^{n}}{\text{min}} d(s^{a}, s^{n})$,  within the  training batch. 
To dynamically adjust sample pairs' weights, we  design  a  novel weight control item.  
\textbf{(i:)}  We introduce the ``nuclear weight  term'',  $\delta=\dfrac{\underset{x^{a},x^{p}}{\text{max}} d(s^{a}, s^{p})}{\underset{x^{a},x^{n}}{\text{min}} d(s^{a}, s^{n})+\varepsilon}$ ($\varepsilon=10^{-6}$). The parameter $\varepsilon$ is  to prevent the denominator from being $0$. 
The  ``nuclear weight  term''   can  adaptively  adjust the weights of  loss  function  under  the hard  positive sample  pairs and  the hard  negative  sample pairs.  When  the  distance  of  hard positive sample pairs becomes larger   or the distance  of  hard negative sample pairs becomes  smaller,   the $\delta$  becomes   larger.  
It indicates that the loss function  pays more  attention to the  current  person ID's hard   sample pairs in the current training  step,  and vice versa. 
\begin{figure}\centering
	\begin{center} 
		\includegraphics[scale=0.33]{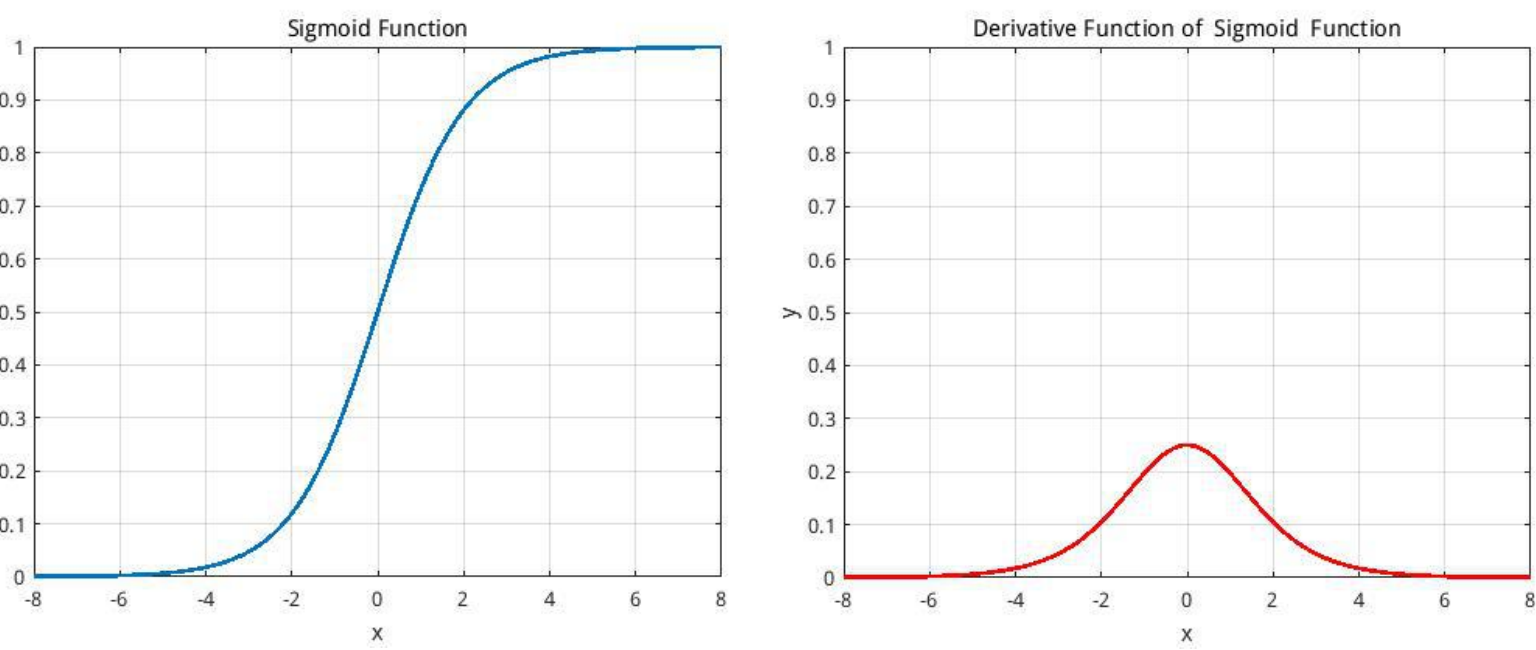}
	\end{center} 
	\caption{ The function  curve  and  derivative  function curve  of  the sigmoid   function. As $x$ gets bigger, the value of  the  sigmoid function  gets closer and closer to $1$. As $x$ gets gets closer and closer to $0$, the value of  the  sigmoid function  gets closer and closer to $\dfrac{1}{2}$.  It can effectively  control the   weight  of  the  Batch  Hard  Triplet  Loss  in $[\dfrac{1}{2}, 1)$.  } 
	\label{tcsvt-4} 
\end{figure}

In  the ideal training goal, the  $\underset{x^{a},x^{p}}{\text{max}} d(s^{a}, s^{p})$ is  smaller  than the    $\underset{x^{a},x^{n}}{\text{min}} d(s^{a}, s^{n})$. 
However,  in fact,  we  observe that  the $\underset{x^{a},x^{p}}{\text{max}} d(s^{a}, s^{p})$ is  usually  larger than the  $\underset{x^{a},x^{n}}{\text{min}} d(s^{a}, s^{n})$ for many person ID  sample pairs.
 If  we  directly use   the $\delta$ to adjust  the weight  of  the  Batch  Hard  Triplet  Loss,  the $\delta$  brings  too much weight fluctuation, which is not conducive to model training.  
Therefore,  we  hope  the   weights of $\delta$ can fluctuate over a small range.   
\textbf{(ii:)} Based  on the  $\delta$,  we  design  a ``Shell function'', i.e. $g(x)=\dfrac{1}{1+e^{-x}}$,  which  is  the sigmoid   function. 
It can effectively  control the   $\delta$  in $[\dfrac{1}{2}, 1)$.   In this way, the weight will not fluctuate too much.  And,  the easy sample  pairs  can also get appropriate weights to participate in the effective training of the person Re-ID model.

For  the sigmoid  function,  we can see  the  function  curve  and  the derivative  function curve  in  Figure~\ref{tcsvt-4}.   As $x$, i.e.  $\delta$ in our weight control item,  gets bigger, the value of  the  sigmoid function  gets closer and closer to $1$.    And  the  value  of  the derivative  function is  smaller. It  indicates   that  the weight of the difficult sample is stable  at around $1$.  In the process of the parameter optimization  process, the gradient  of the Batch  Hard  Triplet  Loss  can gain a large  learning  weight.  As $x$, i.e.  $\delta$ in our weight control item,  gets   closer and closer to $0$, the value of  the  sigmoid function  gets closer and closer to $\dfrac{1}{2}$.     When $\delta$  is    around $0$, the  value  of  the derivative  function is  large.   So,  for  the easy sample pairs,  the loss function  gives  a   flexible and small  weight  to focus on  them.

Based  on the  ``nuclear weight  term'' and ``Shell function'',  our final  weight control item is   $\dfrac{1}{1+e^{-\delta}}, \delta=\dfrac{\underset{x^{a},x^{p}}{\text{max}} d(s^{a}, s^{p})}{\underset{x^{a},x^{n}}{\text{min}} d(s^{a}, s^{n})+\varepsilon}$. We  introduce  the   weight control item  into the  Batch  Hard  Triplet  Loss  to help  the  Re-ID  model  better  foucs  on hard  sample pairs.   Besides, as  described  in ~\cite{Alexander2017}, in person Re-ID  task,  it can be beneficial to pull together samples
from the same class as much as possible~\cite{Chengtri2016, ZhangL}.  Inspired  by ~\cite{Alexander2017},  we  use   the  softplus function $ln(1+e^x)$,  the  Batch  Hard  Triplet  Loss~\cite{Alexander2017}, and our  proposed  weight control item  to  build  our   Elastic Loss, i.e.

\begin{small}
\begin{equation}~\label{eq7}
	\mathcal{T}_{Elastic}= ln(1+ exp(\dfrac{1}{1+e^{-\delta}} (\underset{x^{a},x^{p}}{\text{max}} d(s^{a}, s^{p})- \underset{x^{a},x^{n}}{\text{min}} d(s^{a}, s^{n})))) \\
\end{equation}
\end{small}

Where  $\delta=\dfrac{\underset{x^{a},x^{p}}{\text{max}} d(s^{a}, s^{p})}{\underset{x^{a},x^{n}}{\text{min}} d(s^{a}, s^{n})+\varepsilon}$ and  $\varepsilon=10^{-6}$.   Now, we extend our   Elastic Loss to the whole triplet units in our CBDB-Net, which can be formulated as follows:

\begin{footnotesize}
	\begin{equation} ~\label{eq9}
	\mathcal{L}_{Elastic}(X) = \dfrac{1}{|X|}\sum_{(x^{a},x^{p},x^{n}) \in X}\sum_{i=1}^{m}\mathcal{T}_{Elastic}(\varphi(x^{a})_i,\varphi(x^{p})_i,\varphi(x^{n})_i)
	\end{equation}
\end{footnotesize}
where $|X|$ indicates the number of   triplet units in each  training batch.

\subsection{Network Architecture Overview}~\label{NAO}

In this  subsection, we  revisit  the network architecture and summary loss functions of our CBDB-Net in  the training   stage. 

The overall pipeline of  our CBDB-Net is illustrated  in Figure~\ref{tcsvt-2} in  the training  stage. 
The Backbone Network firstly takes a person image $x$ as input, and then outputs the feature map $T(x) \in \mathbb{R}^{24 \times 8 \times 2048}$. 
Secondly, the  feature  $T(x)$  is fed  into  Consecutive Batch DropBlock  Module (CBDBM). The CBDBM can output multiple  incomplete  feature  descriptors $\varphi(x)_i \in \mathbb{R}^{512},  \emph{i} = 1,2,\cdots, m$.  Finally,  the Cross-Entropy loss  and the Elastic Loss are also  employed at last.  The  whole loss functions  in  training  stage  are listed  as:

\begin{equation}~\label{eq9}
\mathcal{L}_{all}(X)=\mathcal{L}_{Elastic}(X)+ \sum_{j=1}^{M} \sum_{i=1}^{m} \mathcal{L}_{CE}(\varphi(x)_{ij}))
\end{equation}
Here, $\varphi(x)_i \in \mathbb{R}^{512},  \emph{i} = 1,2,\cdots, m$;  $\mathcal{L}_{CE}(\cdot)$ is the Cross-Entropy Loss  for  person ID classification.   The batch size $M$ in the  training stage is $64$.

\section{Experiment}
In this section, we evaluate the CBDB-Net qualitatively  and quantitatively.  To evaluate the effectiveness of our CBDB-Net, we conduct extensive experiments on three generic  person  datasets (the Market-1501~\cite{Zheng2015Scalable}, the DukeMTMC-reID~\cite{Ristani2016Performance, ZhedongZheng},  and  the CUHK03~\cite{Weireid2014}), three occluded Person Re-ID  datasets (the Occluded-DukeMTMC~\cite{Jiaxu112019}, the  Partial-REID~\cite{WeiShi2015}, and  the  Partial-iLIDS~\cite{Lingxiao2018})  and one  clothes  image retrieval dataset (In-Shop Clothes Retrieval dataset~\cite{Ziwei2016}).
Firstly, we discuss various ablation studies on the  four  datasets (the Market-1501~\cite{Zheng2015Scalable}, the DukeMTMC-reID~\cite{Ristani2016Performance, ZhedongZheng},   the CUHK03~\cite{Weireid2014}, and the In-Shop Clothes Retrieval dataset~\cite{Ziwei2016})  to validate the effectiveness of each strategy  in our CBDB-Net.
Secondly, we compare the performance of CBDB-Net against many state-of-the-art methods on these  seven  datasets.

\subsection{Datasets and Evaluation}

\textbf{Market-1501~\cite{Zheng2015Scalable}} contains $32,668$ labeled images of $1,501$ identities which is collected from $6$  different camera views. 
Following almost person Re-ID  approaches,  the  whole $1,501$ identities  are split into two non-overlapping fixed person ID sets: the training set contains $12,936$ person  images from $751$ identities; the testing set contains $19,732$  person  images from other $750$ identities. In the testing stage, we  use $3368$ query images from $750$  test  person  identities  to retrieval the same  ID  persons from the   rest of the test set, i.e. the gallery set. 

\textbf{DukeMTMC-reID~\cite{Ristani2016Performance, ZhedongZheng}}  is also a large-scale person  Re-ID dataset. The DukeMTMC-reID  contains $36,411$ labeled images of $1,404$ identities which is collected from $8$  different camera views. The training set  contains $16,522$  person  images from $702$ identities; In the testing stage, we use $2,228$ query images from the other $702$ identities, and $17,661$ gallery images.

\textbf{CUHK03~\cite{Weireid2014}} is the most challenging of these three  generic   person Re-ID  datasets.  It  composed of $14, 096$ images of $1,467$ identities captured from $6$ cameras.  It provides bounding boxes detected from manual labeling and deformable part models (DPMs),  the latter type is more challenging due to severe bounding box misalignment and cluttered background. Following~\cite{Zhun2020, Kaiyang19, Zuozhuo19, Yifan2018},  we use the $767/700$ split~\cite{Weireid2014}   with the detected images.

\textbf{Occluded-DukeMTMC~\cite{Jiaxu112019}} contains $15,618$ training images, $17,661$ gallery images, and $2,210$ occluded query images.  The   Occluded-DukeMTMC  is introduced  by ~\cite{Jiaxu112019}.  We  use  this dataset to   demonstrate that  our CBDB-Net  also can  achieve   good   performance  on the occluded Person Re-ID task.

\textbf{Partial-REID~\cite{WeiShi2015}}  is a specially designed partial person  Re-ID dataset which  contains $600$ images from $60$ person identities.  And  each person has   $5$ partial images in the query set  and  $5$ full-body images in the gallery set.   These images are collected at a university campus from different viewpoints, backgrounds, and different types of severe occlusion.

\textbf{Partial-iLIDS~\cite{Lingxiao2018}}  is a simulated partial person  Re-ID dataset based on the iLIDS dataset.  It  has a total of $476$ images of $119$ person identities.

\textbf{In-shop clothes retrieval~\cite{Ziwei2016}} is  a  clothes  image  retrieval  dataset.   It  contains $11,735$ classes of clothing items with $54,642$ images. The training set contains $25,882$ images from $3,997$ classes; the testing set contains $28,760$ images from $3,985$ classes. The test set is divided into the query set of $3,985$ classes ($14,218$ images) and the gallery set of $3,985$ classes ($12,612$ images).  We  apply our  CBDB-Net  on  the clothes  image retrieval task  to evidence that  our CBDB-Net can be suitable  for other  image  retrieval  tasks.

\textbf{Evaluation Protocol.} We employ two standard metrics as in most person Re-ID approaches, namely  the mean Average Precision (mAP)   and the cumulative matching curve (CMC) used for generating ranking accuracy. We  use  Rank-1 accuracy and mAP  to evaluate  the effectiveness  of our CBDB-Net   on all seven datasets.

\subsection{Implementation Details}

Following  many outstanding approaches~\cite{Wenjie19, Zuozhuo19, ZhengWang2020, AANET19,  Jiaxu112019, Zhong_2019}, the input images are re-sized to $384 \times 128$ and then augmented by random horizontal flip and  normalization  in the training stage.  In the testing stage, the images are also re-sized to $384 \times 128$ and augmented only by normalization.  Based on the pre-trained ResNet-50 backbone, our network is end-to-end in the whole  training stage.   Our network is trained using  $2$ single GTX $2080$Ti  GPUs with a batch size of $64$. Each batch contains $16$ identities, with $4$ samples per identity.    We use the Adam optimizer~\cite{2015Diederik} with $400$ epochs.  The base learning rate  is  initialized to $1e-3$ with a linear warm-up~\cite{PriyaGoyal2017} in the first $50$ epochs, then decayed to $1e-4$ after $200$ epochs, and further decayed to $1e-5$ after $300$ epochs.

\subsection{Ablation Study of CBDB-Net}

\begin{table}[!t]
	\centering
	\caption{Results of  CBDB-Net  on   the  CUHK03-Detected  and  Market-1501 datasets  under different   number  of  branches in the propopsed  CBDB-Net. }
	\setlength{\tabcolsep}{4mm}{
		\begin{tabular}{c|c c | c c }
			\hline \hline
			\multirow{2}{*}{Method}   &
			
			\multicolumn{2}{|c}{CUHK03-Detected}  & \multicolumn{2}{|c}{Market-1501}    \\
			\cline{2-5} 	
			&  Rank-1 & mAP 	&  Rank-1 & mAP    \\	
			\hline
			m=2   &   58.4\%   &  56.2\%       &   89.0\%   &  74.6\%   \\
			m=3   &   74.2\%   &  69.2\%     &   93.6\%   &  83.8\%    \\
			m=4    &   75.7\%    & 71.1\%     &   94.0\%   &  85.1\% \\
			m=5    &   75.9\%    &  71.9\%     &   94.1\%   &  84.6\%  \\ 
			m=6   &   76.6\%    &  72.8\%     &   94.4\%   &  85.0\%   \\ 
			m=7    &   76.3\%   &  73.0\%      &   94.3\%   &  85.0\%   \\ 	
			m=8   &   76.2\%   &  72.6\%     &   94.3\%   &  84.8\%    \\
			m=9   &   76.4\%   &  72.6\%     &   94.4\%   &  84.9\%   \\
			m=10    &   76.5\%    & 72.7\%   &   94.3\%   &  84.7\%  \\
			m=11   &   74.6\%    &  70.4\%   &   93.9\%   &  84.0\%   \\ 
			m=12  &   75.9\%    &  72.5\%    &   94.3\%   &  84.4\%   \\ 
			\hline  \hline
	\end{tabular} }
	\label{tablex} 
\end{table}

\subsubsection{Discussion of   the number of DropPatches  $m$}~\label{AbTotal2}

\begin{figure}\centering
	\begin{center} 
		\includegraphics[scale=0.15]{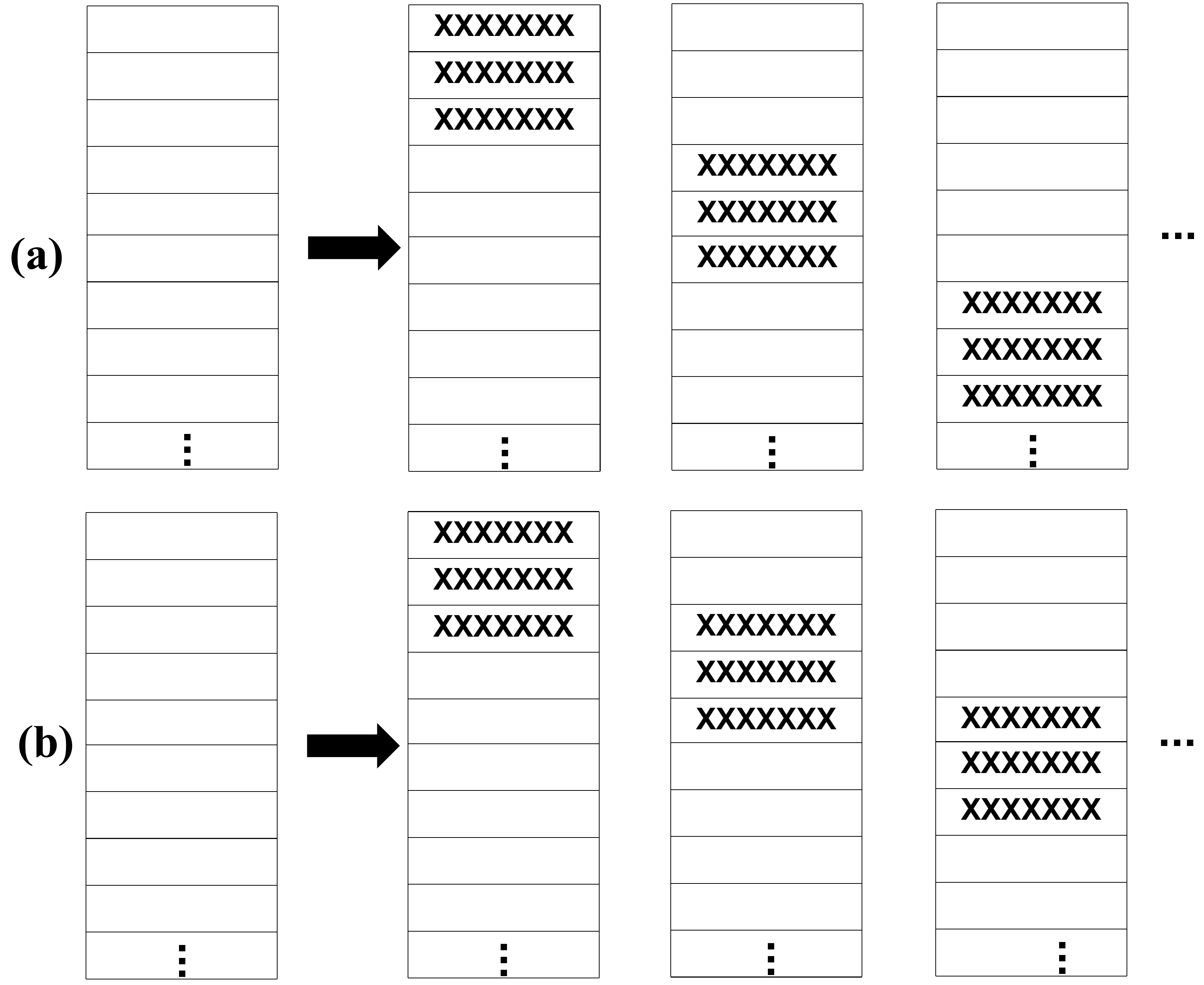}
	\end{center} 
	\caption{Various versions  of   the  Consecutive Batch DropBlock  strategy.  The  figure (a) is the   Consecutive Batch DropBlock  strategy in  Figure~\ref{tcsvt-3}.  There is no  overlap  areas  between the dropped  patch  by  $DropPatch-(i)$  and  dropped  patch  by $DropPatch-(i+1)$.  The figure (b) is the  other Consecutive Batch DropBlock  strategy.  There is an  overlap  area  between the dropped  patch  by  $DropPatch-(i)$  and  dropped  patch  by $DropPatch-(i+1)$.  } 
	\label{tcsvt-7} 
\end{figure}

In  this  subsection,   we  mainly  discuss  the influence  of   the number of DropPatches  $m$  on the  performance  of the  CBDB-Net.  In  our  CBDB-Net, the  size  of $T(x)$   is $24 \times 8 \times 2048$.  When  $m= 2, 3,4, 6, 8,  12$, $24 \% m =0$.  When  $m = 5, 7, 9, 10 ,11$,  $24 \%  m \neq 0$.  In the  Figure~\ref{tcsvt-7}, there are  two kind  Consecutive Batch DropBlock  strategies: in  Figure~\ref{tcsvt-7} (a),   there  is  no   overlap    between the dropped  patch  by  $DropPatch-(i)$  and  dropped  patch  by $DropPatch-(i+1)$;  in  Figure~\ref{tcsvt-7} (b),  there  is  an overlap  area  between the dropped  patch  by  $DropPatch-(i)$  and  dropped  patch  by $DropPatch-(i+1)$.  When  $24 \% m =0$, we  adopt the dropout pattern  of Figure~\ref{tcsvt-7} (a)   to build  our  CBDB-Net. When   $24 \%  m \neq 0$, we  adopt the dropout pattern  of Figure~\ref{tcsvt-7} (b)   to build  our  CBDB-Net.    As  shown  in Table~\ref{tablex}, we  show  the results  of our  CBDB-Net  under  the different  value $m$ in the  CUHK03-Detected  and Market-1501 datasets. When $m$ is equal to $6$,  our CBDB-Net  achieves  the best  performance  on these two  datasets.    So,  in our CBDB-Net, we  set  the number of DropPatches is $6$, i.e. $m=6$.

\subsubsection{Effectiveness  of the each strategy in the  CBDB-Net}~\label{AbTotal1}

\begin{table*}[!t]
	\centering
	\caption{Results produced by combining different components of the  CBDB-Net.  The parameter $m$ in  this  table  is  $6$. CBDB-Net$^{\dagger}$:  we  directly  concatenate   the  $\varphi(x)_i \in \mathbb{R}^{512},  \emph{i} = 1,2,\cdots, 6$  to  conduct  the  person matching  task in  the  testing stage.   IDE+CBDB-Net contains $m+1$ branches:  $m$ incomplete  feature   branches   and a  global branch from IDE~\cite{Zuozhuo19, Zheng2016Person}. The training pipeline  and testing pipeline  of the IDE+CBDB-Net  are  shown  in Figure~\ref{tcsvt-8} and Figure~\ref{tcsvt-9}  respectively.   CBDBM$^*$ = CBDB-Net w/o the Elastic Loss = IDE+CBDBM w/o the global branch from IDE~\cite{Zuozhuo19, Zheng2016Person}.        BHT is short for Batch Hard Triple Loss.  EL is short for Elastic Loss. The Bold is the best result.  }
	
	\setlength{\tabcolsep}{2.5mm}{
		\begin{tabular}{c|c c | c c | c c | c c| c c }
			\hline \hline
			\multirow{2}{*}{Method}   &
			
			\multicolumn{2}{|c|}{Market-1501} &\multicolumn{2}{|c}{DukeMTMC-reID} &\multicolumn{2}{|c}{CUHK03-Detected} &\multicolumn{2}{|c}{CUHK03-Labeled}  &\multicolumn{2}{|c}{Clothes}  \\
			\cline{2-11} 	
			&  Rank-1 & mAP & Rank-1 & mAP & Rank-1 & mAP & Rank-1 & mAP  & Rank-1 & mAP \\	
			\hline
			IDE+BHT~\cite{Zuozhuo19, Zheng2016Person}    &   93.1\%   & 80.6\%   &   84.4\%   & 68.4\%  &   63.6\%   & 60.0\%  &   67.4\%   & 61.5\%    & 89.9\% & 72.0\% \\  \hline  \hline  
			IDE+EL    & 93.9\% &   82.5\%       &    85.9\%  & 71.2\%     &    69.9\% & 65.9\%        &  70.3\% &  66.5\%     & 91.7\% & 73.8\% \\
				IDE+CBDBM   & 94.2\% &   84.4\%       &    87.3\%   & 73.6\%     &    75.3\% & 71.8\%        &  77.6\% &  74.7\%    & 91.2\% & 73.6\% \\
			CBDBM$^*$     &  94.0\% &   84.7\%    &    86.5\%  & 70.7\%   &    74.7\%    &  71.5\%      & 75.7\%  & 73.2\%  & 91.0\% & 73.1\% \\
			CBDB-Net     & \textbf{94.4\%} &   85.0\%       &    87.7\%   &  74.3\%     &    \textbf{76.6\%} & 72.8\%        &  \textbf{78.3\%} &  \textbf{75.9\%}     & 92.0\%  & 75.4\% \\
			 \hline \hline
			CBDB-Net w/o ``ResBlock''     & 93.8\% &   83.7\%      &   86.5\%   &  72.2\%  &    73.4\%  & 69.5\%       &  76.6\% &  73.0\%   & 91.8\% & 74.4\% \\
		
			CBDB-Net$^{\dagger}$     & \textbf{94.4\%} &   \textbf{85.2\%}      &    87.3\%   &  73.9\%   &    74.9\%  & 72.0\%       &  77.8\% &  75.5\%   & 92.1\%  & 75.6\%  \\
			IDE+CBDB-Net  & 94.3\% &   85.0\%       &    \textbf{88.0\%}   &  \textbf{74.4\%}     &    \textbf{76.6\%} & \textbf{73.1\%}        & 77.8\% &  75.4\%    & \textbf{92.2\%} & \textbf{75.6\%}  \\
			\hline
	\end{tabular} }
	\label{table5} 
\end{table*}

\begin{figure}\centering
	\begin{center} 
		\includegraphics[scale=0.14]{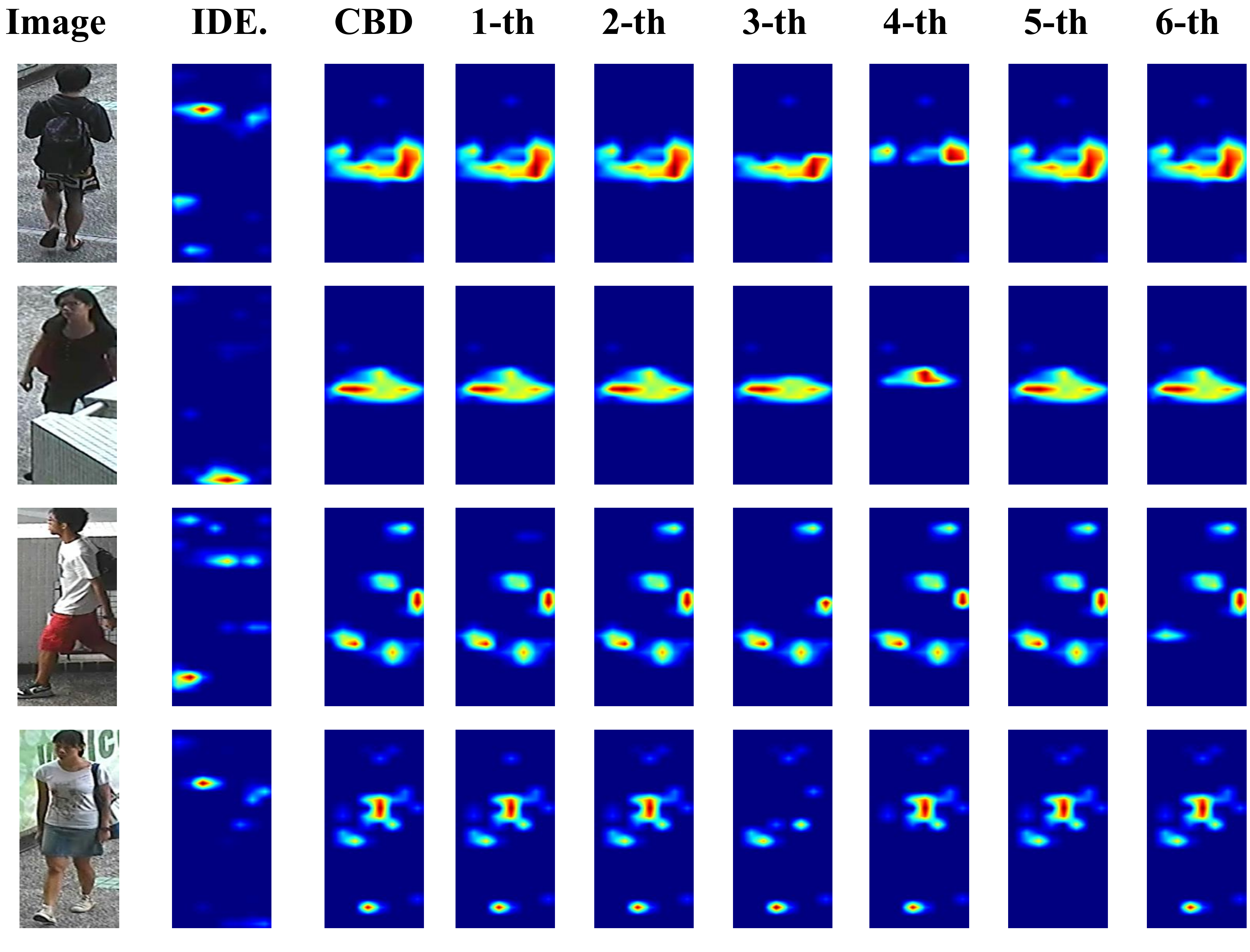}
	\end{center} 
	\caption{Visualization of attention maps from Baseline, CDBD-Net  and  six branches (m=$6$) of  the CBDB-Net.   The attention maps  of Baseline  are   in the  $1-st$ and $2-nd$ column; The attention maps  of  our CBDB-Net  are in the $3-rd$ column;   The attention maps  of six branches (m=$6$) of  the CBDB-Net are from the $4-th$ column to the $9-th$ column respectively. } 
	\label{tcsvt-6} 
\end{figure}

In  this  subsection,  we  mainly discuss  the  effectiveness  of  each   component  in our CBDB-Net  on   four  datasets: Market-1501,  DukeMTMC-reID,  CUHK03-Detected,   and  In-shop clothes retrieval datasets.

\textbf{(I:)} To demonstrate  the  effectiveness of  the  proposed Elastic Loss,  we  use  the IDE+BHT\cite{Zuozhuo19, Zheng2016Person}  as  the  baseline  model.  Here, BHT is short for Batch Hard Triple Loss. 
Based  on the IDE+BHT,    we replace  Batch  Hard  Triplet Loss    with the  proposed  Elastic Loss, i.e. ``IDE+EL''.  Compared with IDE+BHT, our IDE+EL gains the  obvious  improvements  on the four  datasets  over two  measures.  It effectively  evidences the effectiveness  of   our Elastic Loss.

\begin{figure}\centering
	\begin{center} 
		\includegraphics[scale=0.32]{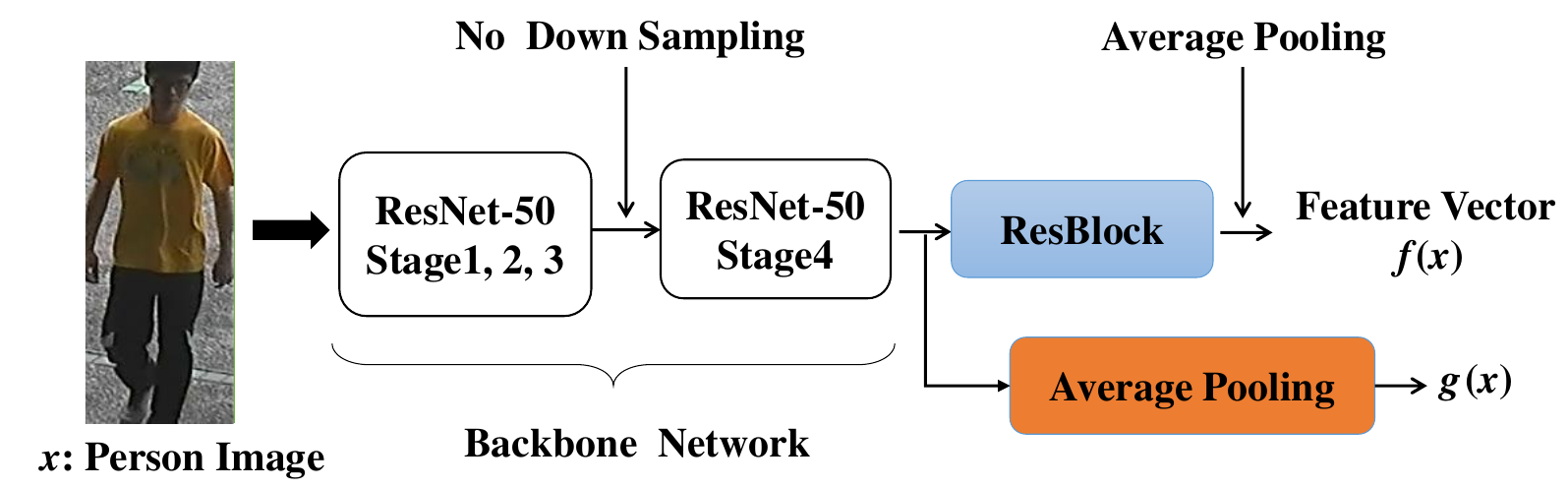}
	\end{center} 
	\caption{The network  pipeline of the IDE+CBDB-Net  or IDE+CBDBM   in the testing stage. The concatenation between vector $f(x)$  and $g(x)$ is  the feature  vector  in the testing stage. } 
	\label{tcsvt-9} 
\end{figure}

\textbf{(II:)} To demonstrate the effectiveness of  the  Consecutive Batch DropBlock  Module (CBDBM),  we also  use  the IDE+BHT\cite{Zuozhuo19, Zheng2016Person} as  the  baseline  model.  We demonstrate the validity of the CBDBM in two similar ways. 
\textbf{(a:)} We  introduce  the CBDBM  into  IDE+BHT\cite{Zuozhuo19, Zheng2016Person}, i.e. IDE+CBDBM.  
Compared with IDE+BHT, our IDE+CBDBM  also  gains the  obvious  improvements in the four  datasets  over two  measures.  Besides, in  the  ``IDE+CBDBM'',   the testing  feature  vector   is  composed  of    the feature  vector $f(x)$ (Dimension=512) and  a  global  feature  vector $g(x)$ (Dimension=512)  from the global  branch in  IDE~\cite{Zuozhuo19, Zheng2016Person}.  The  testing  pipeline  of  the   ``IDE+CBDBM''   can be  regarded  as the pipeline  shown in Figure~\ref{tcsvt-9}. 
\textbf{(b:)} We drop out  the  global  feature  branch  in the  ``IDE+CBDBM'', i.e.  ``CBDBM$^*$''. Here, CBDBM$^*$ = CBDB-Net w/o the Elastic Loss = IDE+CBDBM w/o the global branch from IDE~\cite{Zuozhuo19, Zheng2016Person}.  As  shown  in Table~\ref{table5},  compared with IDE+BHT, our  CBDBM$^*$  also  gains the  obvious  improvements  on the four  datasets  over two  measures. Besides, from the $4-th$ column to the $9-th$ column  in Figure~\ref{tcsvt-6}, each  branch  in CBDBM  can  also capture the key  information from  the  various incomplete  feature  maps.

\textbf{(III:)} Our  CBDB-Net  contains two  novel  designs: the  Consecutive Batch DropBlock  Module (CBDBM)  and  the Elastic Loss.   As shown in Table~\ref{table5}, compared  with IDE+BHT\cite{Zuozhuo19, Zheng2016Person},  our CBDB-Net achieves better  performance.  Besides, as  shown  in Figure~\ref{tcsvt-6},  compared  with Baseline (IDE+BHT~\cite{Zheng2016Person}), our  CBDB-Net  can capture  more  key person information.   \textbf{The CBDB-Net  is  our  proposed  novel  person Re-ID model  in this  paper. }  The  CBDB-Net's   training model  is  shown  in  Figure~\ref{tcsvt-2}, and   CBDB-Net's   testing  model  is  shown  in  Figure~\ref{tcsvt-5}.
In the  testing stage, we use the $512$  dimension  feature vector  $f(x)$  to  conduct   the  person matching  task. 

\begin{figure}\centering
	\begin{center} 
		\includegraphics[scale=0.17]{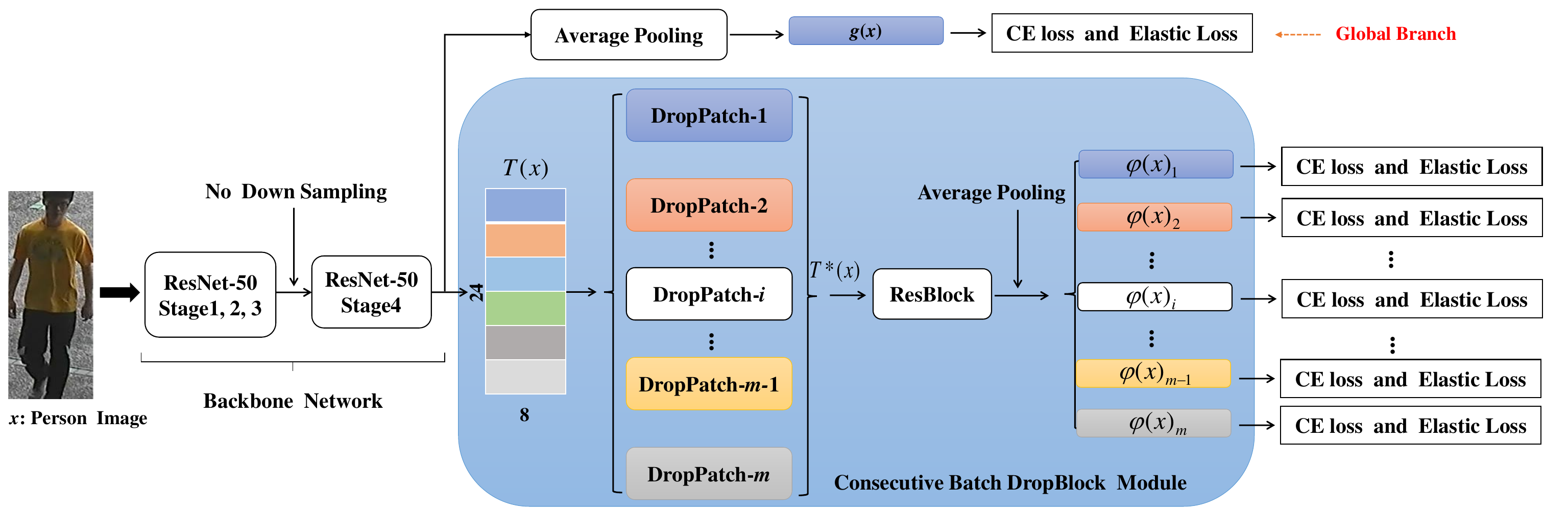}
	\end{center} 
	\caption{The architecture of the IDE+CBDB-Net for the person Re-ID task.  The IDE~\cite{Zuozhuo19, Zheng2016Person} contains the Backbone Network, Average Pooling,  	CE  Loss  and  Elastic Loss.  The ``CE'' denotes the  Cross-Entropy  Loss function.   } 
	\label{tcsvt-8} 
\end{figure}

\textbf{(IV:)} In  our  CBDB-Net, based  on the  CBDBM, we can gain  multiple incomplete  feature  maps. As  shown in  Figure~\ref{tcsvt-2}, we  append the additional   ``ResBlock''  and average pooling operation  on  these  incomplete  feature tensors to  gain  incomplete   person descriptors.  
To evidence the  influence of  the  ``ResBlock'',  we   directly append  the average  pooling  operation  on these  incomplete  feature tensor $T^*(x)$ to conduct the person  retrieval task, i.e. CBDB-Net w/o ``ResBlock''. The  results  in Table~\ref{table5} show that   the  ``ResBlock'' also plays an important  role in the person matching task.

\textbf{(V:)}  We further  concatenate  the  $\varphi(x)_i \in \mathbb{R}^{512},  \emph{i} = 1,2,\cdots, 6$, CBDB-Net$^{\dagger}$,   and then  use  $6 \times 512$  dimension  feature vector  to  conduct  the person  matching.
As  shown  in Table~\ref{table5},  the CBDB-Net$^{\dagger}$  can  not  achieve a  better  performance  than  that  of  the CBDB-Net. 
Even on the Duke-MTMC, CUHK03-Detected, and CUHK03-Labeled datasets,  the performance  of the CBDB-Net$^{\dagger}$  is worse than that of  the CBDB-Net.  And  compared with CBDB-Net, the dimension of the  feature vector of  the CBDB-Net$^{\dagger}$ in the testing stage  is much  higher  than that of the CBDB-Net.    It  indicates  that the feature $f(x)$  of   CBDB-Net  is a   high  quality  and robust global  descriptor  for the person matching task.

\textbf{(VI:)} Our  CBDB-Net  does  not contain   the  global feature  vector  from the  global  branch of  IDE~\cite{Zheng2016Person}. So,  we  introduce  the global  branch into  the CBDB-Net, i.e. IDE+CBDB-Net.   The training pipeline  and testing pipeline  of the IDE+CBDB-Net  are  shown  in Figure~\ref{tcsvt-8} and Figure~\ref{tcsvt-9} respectively.   The concatenation between vector $f(x)$  and $g(x)$ is  the feature  vector  in the testing stage.  Compared  with CBDB-Net,  the IDE+CBDB-Net   only   gains a slight  improvement.  It indicates that the additional  global branch does not make an effective contribution to the improvement of the person Re-ID task.  To keep the model simple, we  select  the CBDB-Net as  our  proposed  novel  person Re-ID  model  in this  paper.

\subsubsection{Comparison with  Various   Dropout Strategies}~\label{AbTotal3}

In   this  subsection,  we  discuss  the performance of  various   Dropout  strategies on the  CUHK03-Detected  dataset. 
\textbf{(i:)}  Dropout~\cite{Nitish2014} randomly  drops the neural node of input  feature maps  or  vector,  which is  regarded  as  a  kind of  regularization strategy to prevent  overfitting of  deep  models.
\textbf{(ii:)} Compared  with Dropout,  SpatialDropout~\cite{Jonathan2015} further  randomly zeroes whole channels of the input feature  tensors. 
And  the channels of feature  tensors  to zero-out are randomized.  
\textbf{(iii:)}  Based  on the SpatialDropout,  Batch Dropout randomly  zeroes  the  whole channels of the input feature  tensors within the  whole batch.  
However, the Batch Drop can  not drop   a large contiguous area.  
\textbf{(iv:)} So, the DropBlock~\cite{Golnaz2018}   strategy  randomly drops a contiguous region on the feature maps. 
\textbf{(v:)} the Batch DropBlock  further  randomly  drops the same region for every input tensor within a batch. 
Different from the Batch DropBlock~\cite{Zuozhuo19},   our  Consecutive Batch  DropBlock  can    drop  the same  region for every input tensor within  the  whole  training set  and product multiple  incomplete feature tensors.    
As  shown in Table~\ref{table8}, compared with  these  Dropout strategies,     our  Consecutive  Batch DropBlock  strategy  achieves the best performance  on the CUHK03-Detected  dataset over  two  measures.

\begin{table}[!t]
	\centering
	\caption{The Comparison with 	a  series  of  Dropout strategies on the	CUHK03-Detected dataset.  The Bold is the best result.}
	
	\setlength{\tabcolsep}{6mm}{
		\begin{tabular}{c| c c  }
			\hline \hline
			\multirow{2}{*}{Method}   &
			\multicolumn{2}{|c}{CUHK03-Detected}  \\
			\cline{2-3} 	
			&   Rank-1 &  mAP   \\	
			\hline
			SpatialDropout~\cite{Jonathan2015}     & 60.5\% &  56.8\%     \\ 
			Dropout~\cite{Nitish2014}    & 65.3\%  &  62.2\%     \\ 
			Batch Dropout~\cite{Zuozhuo19}     & 65.8\%  & 62.9\%     \\ 
			DropBlock~\cite{Golnaz2018} &   70.6\%  & 67.7\% \\
			Batch DropBlock~\cite{Zuozhuo19} &  72.8\%  &   69.3\%  \\ \hline
			CBDB &  \textbf{75.3\%}  &  \textbf{71.8\%}  \\ 
			\hline   \hline
	\end{tabular} }
	\label{table8} 
\end{table}

\subsubsection{Discussion of  the  model  size  and   testing time}~\label{AbTotal2}
 
In  this  subsection,  we  mainly  discuss  the  model  size  and   testing time  between  our  CBDB-Net  and  other  SOTA  person Re-ID  models.  As  shown in Table~\ref{model_size}, we  list the  Model  Size, Testing  Vector 	Dimension, and  Testing Time   of our CBDB-Net  and other  SOTA  person Re-ID  models  on the Market-1501 test  set.

\textbf{(I) Model Size.} The Backbone Network of  these  methods  listed in Table~\ref{model_size}  is  ResNet-50.   The released  code  of these  methods  (include our CBDB-Net) is PyTorch.  The model  size listed  in Table~\ref{model_size} is the size of  ``.pth''  document. Compared  with  BDB~\cite{Zuozhuo19}, PCB~\cite{Yifan2018}, PGFA~\cite{Jiaxu112019}, and  ABD-Net~\cite{Tianlong2019},    the model  size  of our CBDB-Net  is the  smallest.

\textbf{(II) Testing  Vector 	Dimension.}  As shown  in  Table~\ref{model_size}, the  testing vector of our  CBDB-Net is  only $512$, which  is  much smaller  than that of BDB~\cite{Zuozhuo19}, PCB~\cite{Yifan2018}, PGFA~\cite{Jiaxu112019}, and  ABD-Net~\cite{Tianlong2019}.

\textbf{(III) Testing Time.}  As shown  in  Table~\ref{model_size}, we  list  the testing time  of  these Re-ID models on the  Market-1501  test  set.   We run the test  experiments  on an Intel Core i$7-7820$  CPU@$3.60Hz\times16$, with $64$GB RAM and an RTX $2080$ Ti GPU.   As  shown in Table~\ref{model_size}, the testing time of our CBDB-Net  is  the  shortest.

Above  all,  our  CBDB-Net  is a simple  and effective person Re-ID  model.

\begin{table}[h]
	\centering
	\caption{ The  model  size, testing  vector 	dimension, and   the testing time   of our CBDB-Net  and other person Re-ID  models  on the Market-1501 dataset. }	
	\setlength{\tabcolsep}{0.5mm}{
		\begin{tabular}{c|c|c | c |c | c }
			\hline	\hline 
			$\textbf{Method}$     & CBDB-Net & BDB~\cite{Zuozhuo19}  &  PCB~\cite{Yifan2018}  & PGFA~\cite{Jiaxu112019} &  ABD-Net~\cite{Tianlong2019}     \\   \hline
			Model size     & \textbf{100M}  & 129.4M  & 109.2M  & 110M  &  264M  \\  
			Dimension   &  \textbf{512} & 1536  & 12288  & 1792 &  2048  \\  
			Testing Time     & \textbf{37s}  & 42s  & 120s  & 172s  &  106s \\  \hline	\hline		
	\end{tabular} }
	\label{model_size}  	
\end{table}

\subsection{Comparison to State-of-the-art Methods}

\begin{table*}[!htbp]
	\centering
	\caption{The comparison with many  state-of-the-art  person re-ID approaches on the  Market-1501,the DukeMTMC-reID and the CUHK03 datasets. The  results  of  the ``CBDB-Net+Re-ranking''   are  bold. In  addition to  the``CBDB-Net+Re-ranking'' , 
		the \textcolor{red}{first}, \textcolor{ForestGreen}{second} and \textcolor{blue}{third} highest scores are shown in \textcolor{red}{red}, \textcolor{ForestGreen}{green} and \textcolor{blue}{blue} respectively. We set $m=6$.}
	
	\setlength{\tabcolsep}{4mm}{
		\begin{tabular}{c| c c | c c |  c  c | c  c }
			\hline \hline
			\multirow{2}{*}{Method}   
			&\multicolumn{2}{|c|}{Market-1501} &\multicolumn{2}{|c}{DukeMTMC-reID} &\multicolumn{2}{|c}{CUHK03-Detected} &\multicolumn{2}{|c}{CUHK03-Labeled}  \\
			\cline{2-9} 	
			& Rank-1 & mAP & Rank-1 & mAP   & Rank-1 & mAP   & Rank-1 & mAP \\	
			\hline
			
			MCAM\cite{Chunfeng2018}    & 83.8\% & 74.3\% & - & -  &  46.7\% &  46.9\%  &50.1\% & 50.2\% \\ 
			MLFN \cite{Xiaobin18}   & 90.0\% & 74.3\% &81.0\% & 62.8\% & 52.8\% & 47.8\%  & 54.7\%  & 49.2\%  \\ 
			SPReID\cite{Mahdi2018cvpr}  &   92.5\% &  81.3\%   &   84.4\%   & 71.0\%  & - & - & - & - \\
			HA-CNN~\cite{li2018harmonious}   &  91.2\%  & 75.7\%  & 80.5\%  & 63.8\%  & 41.7\% & 38.6\%  &  44.4\% & 41.0\% \\
			PCB+RPP\cite{Yifan2018}   & 93.8\%   & 81.6\%    &    83.3\%   &   69.2\%    & 62.8\%   & 56.7\% & - & - \\
			Mancs\cite{cheng2018eccv}     &   93.1\%   & 82.3\%   &   84.9\%   & 71.8\%   &   65.5\%   & 60.5\% \\ 
			JSTL\_DGD+ICV-ECCL ~\cite{Zhanxiang2018}    & 88.4\%   & 69.5\%  & -   & -  & - & - & - & - \\
			PAN~\cite{Zheng_2018}   & 82.8\% & 63.4\%   & 71.6\% & 51.5\%  & 36.3\% & 34.0\% & 36.9\% & 35.0\% \\ 
			Camstyle\cite{Zhong_2019}  & 88.1\%   & 68.7\%  & 75.3\%   & 53.5\%  & - & - & - & - \\
			FANN\cite{Discriminative_2019}   & 90.3\%   & 76.1\%  & -& - & 69.3\%   & 67.2\%   & -& - \\
			VCFL\cite{Fangyi19}     &   90.9\% & 86.7\%      & - & -   &   \textcolor{blue}{70.4\%} & \textcolor{ForestGreen}{70.4\%} &  - &  - \\	
			PGFA~\cite{Jiaxu112019}   &   91.2\%   & 76.8\%  &   82.6\%   & 65.5\%   &   -   & -  & -  & - \\    
			SVDNet+Era~\cite{Zhun2020}    & 87.1\%  &  71.3\%  & 79.3\%  & 62.4\% & 48.7\%  &  37.2\%  & 49.4\% & 45.0\%  \\
			TriNet+Era\cite{Zhun2020}   &  83.9\%  & 68.7\%  &  73.0\% & 56.6\%  &    55.5\%   & 50.7\% & \textcolor{blue}{58.1\%}  & \textcolor{blue}{53.8\%} \\  
			HACNN+DHA-NET~\cite{ZhengWang2020}  & 91.3\%   & 76.0\%  & 81.3\%  & 64.1\%  & - & - & - & - \\		
			IANet\cite{Ruibing19}     &   \textcolor{ForestGreen}{94.4\%}   & 83.1\% &   \textcolor{ForestGreen}{87.1\%}   & \textcolor{ForestGreen}{73.4\%}  & - & -  & - & - \\ 
			BDB\cite{Zuozhuo19}     &   \textcolor{blue}{94.2\%}   & \textcolor{blue}{84.3\%}  &   \textcolor{blue}{86.8\%}   & 72.1\%  &   \textcolor{ForestGreen}{72.8\%}   & \textcolor{blue}{69.3\%} & \textcolor{ForestGreen}{73.6\%} & \textcolor{ForestGreen}{71.7\%}\\
			AANet\cite{AANET19}     &   93.9\%   & 83.4\%   &   \textcolor{red}{87.7\%}   & \textcolor{red}{74.3\%}  & - & - & - & - \\ 
			CAMA\cite{Wenjie19}      &  \textcolor{red}{94.7\%}  & \textcolor{ForestGreen}{84.5\%}  &   85.8\%   & \textcolor{blue}{72.9\%}   &   66.6\%   & 64.2\% \\    \hline
			CBDB-Net           & \textcolor{ForestGreen}{94.4\%} &   \textcolor{red}{85.0\%}       &     \textcolor{red}{87.7\%}   &   \textcolor{red}{74.3\%}     &     \textcolor{red}{76.6\%} &  \textcolor{red}{72.8\%}        &  \textcolor{red}{78.3\%} &  \textcolor{red}{75.9\%}      \\
			CBDB-Net+Re-ranking~\cite{Zhong2017Re}             & \textbf{95.6\%} & \textbf{93.0\%}   &    \textbf{91.2\%} & \textbf{87.9\%}  &    \textbf{83.9\%} & \textbf{85.1\%} &  \textbf{86.5\%} &  \textbf{87.8\%}    \\
			
			\hline \hline
	\end{tabular} }
	\label{table1} 
\end{table*}

Firstly,  we  evaluate   the  performance   of  CBDB-Net  on the  generic person Re-ID task. We compared our  CBDB-Net   against the  many  state-of-the-art  approaches on  Market-1501, DukeMTMC-Re-ID and CUHK03, as shown in Tables~\ref{table1} respectively.  
From the Table~\ref{table1}, we can observe that our CBDB-Net achieves  competitive performance on these  three generic   person  Re-ID datasets and  outperforming most published approaches by a clear margin. 
Specifically,  CBDB-Net obtains $94.4\%$ Rank-1  and $85.0\%$ mAP, which  outperforms most existing methods on Market-1501 dataset. 
And then,  we further  introduce the Re-Ranking~\cite{Zhong2017Re}  into  our  CBDB-Net, i.e.  CBDB-Net+Re-ranking.  
Here,  the   CBDB-Net+Re-ranking  can   achieve  $95.6\%$ Rank-1 and $93.0\%$ mAP  on  the  Market1501.
On the DukeMTMC-reID dataset,  our CBDB-Net obtains $87.7\%$ Rank-1  and $74.3\%$ mAP.  
The CBDB-Net+Re-ranking  can   achieve  $91.2\%$ Rank-1 and $87.9\%$ mAP.  
The CUHK03 dataset is  the most challenging dataset among the three  generic person Re-ID  datasets.   
Following the data setting in ~\cite{Zhun2020, Kaiyang19, Zuozhuo19, Yifan2018}, our CBDB-Net has clearly yielded  good   performance.  
On the CUHK03-Detected dataset, our  CBDB-Net achieves   $76.6\%$ Rank-1 and $72.8\%$ mAP;  On the  CUHK03-Labeled dataset, our  CBDB-Net achieves   $78.3\%$ Rank-1 and $75.9\%$ mAP.  If  we  introduce  the Re-ranking strategy into  the CBDB-Net, the  CBDB-Net+Re-ranking  can further  achieve  $83.9\%$ Rank-1 and $85.1\%$ mAP   on  the   CUHK03-Detected dataset and achieves  $86.5\%$ Rank-1 and $87.8\%$ mAP on the  CUHK03-Labeled dataset  respectively.

\begin{table}[!t]
	\centering
	\caption{ The comparison with a  series  of   occluded  person re-ID methods in Occluded-DukeMTMC dataset.  	The \textcolor{red}{first}, \textcolor{ForestGreen}{second} and \textcolor{blue}{third} highest scores are shown in \textcolor{red}{red}, \textcolor{ForestGreen}{green} and \textcolor{blue}{blue} respectively.  }
	
	\setlength{\tabcolsep}{2.5mm}{
		\begin{tabular}{c| c c c  c}
			\hline \hline
			\multirow{2}{*}{Method}   &
			\multicolumn{4}{|c}{Occluded-DukeMTMC}  \\
			\cline{2-5} 	
			&   Rank-1 & Rank-5 &  Rank-10 &  mAP  \\	
			\hline
			LOMO+XQDA~\cite{Liao2015Person}   & 8.1\% &  17.0\% &  22.0\% & 5.0\% \\ 
			DIM~\cite{2017devil}   & 21.5\% & 36.1\% &  42.8\% &  14.4\% \\
			Part Aligned~\cite{Zhao_2017} &  28.8\%  & 44.6\% & 51.0\% & 20.2\% \\ 
			Random Erasing~\cite{2017Zhong} & 40.5\% & 59.6\% &  66.8\% & 30.0\%\\
			HA-CNN~\cite{li2018harmonious} & 34.4\% & 51.9\%  &  59.4\%  & 26.0\% \\
			Adver Occluded~\cite{2018Adversarially} &  \textcolor{blue}{44.5\%} &  - & - & 32.2\% \\
			PCB~\cite{Yifan2018} &  42.6\%   &  57.1\%  &  62.9\%   & \textcolor{blue}{33.7\%}  \\  
			Part Bilinear~\cite{201xf8Yumin} &  36.9\%   & - &  - &  -  \\
			FD-GAN~\cite{Yixiao2018} &  40.8\%   & - &  - &  -  \\  
			DSR~\cite{Lingxiao2018} & 40.8\%  &  58.2\%  &  65.2\%  & 30.4\%  \\
			SFR~\cite{Lingxiao2018xl} &  42.3\%  &  \textcolor{blue}{60.3\%}   &  \textcolor{blue}{67.3\%}  &  32.0\%  \\
			PGFA~\cite{Jiaxu112019} &  \textcolor{red}{51.4\%}  &  \textcolor{red}{68.6\%}  & \textcolor{red}{74.9\%}  &  \textcolor{ForestGreen}{37.3\%}  \\ \hline
			CBDB-Net &  \textcolor{ForestGreen}{50.9\%} &  \textcolor{ForestGreen}{66.0\%}  & \textcolor{ForestGreen}{74.2\%} & \textcolor{red}{38.9\%}  \\  \hline \hline
	\end{tabular} }
	\label{table2} 
\end{table}

\begin{table}[!t]
	\centering
	\caption{The comparison with a  series  of  occluded    person re-ID methods in Partial-REID  and  Partial iLIDS  dataset. 	The \textcolor{red}{first}, \textcolor{ForestGreen}{second} and \textcolor{blue}{third} highest scores are shown in \textcolor{red}{red}, \textcolor{ForestGreen}{green} and \textcolor{blue}{blue} respectively.}
	\setlength{\tabcolsep}{3mm}{
		\begin{tabular}{c|c c | c c}
			\hline \hline
			\multirow{2}{*}{Method}   &
			
			\multicolumn{2}{|c|}{Partial-REID} &\multicolumn{2}{|c}{Partial iLIDS}  \\
			\cline{2-5} 	
			&  Rank-1 & Rank-3 & Rank-1 & Rank-3  \\	
			\hline
			MTRC~\cite{Shengcai2013}    &   23.7\%   &  27.3\%     &  17.7\%  & 26.1\% \\ 	
			AMC+SWM~\cite{WeiShi2015}   &   37.3\%   &  46.0\%     &  21.0\%  & 32.8\% \\
			DSR~\cite{Lingxiao2018}   &   50.7\%   &  70.0\%     &  58.8\%  & 67.2\% \\
			SFR~\cite{Lingxiao2018xl}    &   \textcolor{blue}{56.9\%}    & \textcolor{ForestGreen}{78.5\%}  &  \textcolor{blue}{63.9\%}  & \textcolor{blue}{74.8\%} \\
			PGFA~\cite{Jiaxu112019}    &   \textcolor{red}{68.0\%}    &  \textcolor{red}{80.0\%}   &  \textcolor{red}{69.1\%} & \textcolor{ForestGreen}{80.9\%} \\ 	\hline 
			CBDB-Net   &   \textcolor{ForestGreen}{66.7\%}    &  \textcolor{blue}{78.3\%}   & \textcolor{ForestGreen}{68.4\%} & \textcolor{red}{81.5\%} \\ 
		 \hline  \hline
	\end{tabular} }
	\label{table3} 
\end{table}

\begin{table}[!t]
	\centering
	\caption{The comparison  on  Rank-1, Rank-10, and Rank-20  with other  methods Stanford online products datasets. 	The \textcolor{red}{first}, \textcolor{ForestGreen}{second} and \textcolor{blue}{third} highest scores are shown in \textcolor{red}{red}, \textcolor{ForestGreen}{green} and \textcolor{blue}{blue} respectively.}
	
	\setlength{\tabcolsep}{2.6mm}{
		\begin{tabular}{c| c c c }
			\hline \hline
			\multirow{2}{*}{Method}   &
			\multicolumn{3}{|c}{In-Shop Clothes}  \\
			\cline{2-4} 	
			&   Rank-1 &  Rank-10  &  Rank-20  \\	
			\hline
			FasionNet~\cite{Ziwei2016}      &  53.0\%  & 73.0\%   &   76.0\%   \\ 
			HDC~\cite{Yuhui2017}       &  62.1\%  & 84.9\%   &   89.0\%   \\  
			DREML~\cite{Hong2018}      &  78.4\%  & 93.7\%   &   95.8\%   \\
			HTL~\cite{Weifeng2018}      &  80.9\%  & 94.3\%   &   95.8\%   \\
			A-BIER~\cite{Michael2018}      &  83.1\%  & 95.1\%   &   96.9\%   \\  
			ABE-8~\cite{Wonsik2018}      &  \textcolor{blue}{87.3\%}  & \textcolor{ForestGreen}{96.7\%}   &   \textcolor{ForestGreen}{97.9\%}   \\
			BDB~\cite{Zuozhuo19}     &  \textcolor{ForestGreen}{89.1\%}  & \textcolor{blue}{96.3\%}   &   \textcolor{blue}{97.6\%}   \\  \hline
			CBDB-Net  &  \textcolor{red}{92.3 $\pm$ 0.3\%}  & \textcolor{red}{98.4 $\pm$ 0.2\%}   &   \textcolor{red}{99.2 $\pm$ 0.2\%}   \\ 
			
			\hline   \hline
	\end{tabular} }
	\label{table4} 
\end{table}

In our CBDB-Net, the  Consecutive DropBlock  Module can produce  multiple incomplete  feature maps.  We can regard the  incomplete  feature map as  a kind of  occluded  or partial person feature.  
As  shown  in  Figure~\ref{tcsvt-6}, these  incomplete  feature tensors  push  the  model to  capture a robust person  descriptor from  the incomplete  feature map for  Re-ID  in the training stage. 
So,  we secondly try to  evaluate   the  performance   of our CBDB-Net  in the  occluded or partial person Re-ID task.  
We compared our  CBDB-Net   against the   many  approaches on  Occluded DukeMTMC, Partial-REID,  and Partial iLIDS,  as shown in Table~\ref{table2} and  Table~\ref{table3}  respectively.   
On  the Occluded DukeMTMC  dataset,  our  CBDB-Net  achieves the   $50.9\%$ Rank-1 and $38.9\%$ mAP;  On  the Partial-REID  dataset,  our  CBDB-Net  achieves the   $66.7\%$ Rank-1 and $78.3\%$ Rank-3;  On  the Partial iLIDS dataset,  our  CBDB-Net  achieves the   $68.4\%$ Rank-1 and $81.5\%$ Rank-3. 
From  the  Table~\ref{table2}  to   Table~\ref{table3}, on the  occluded or partial  person Re-ID task,  our  CBDB-Net  achieves  the  competitive results  on  the  three datasets. Compared  with PGFA~\cite{Jiaxu112019}, the performance of  our CBDB-Net is a little bit worse. 
However,   the  PGFA~\cite{Jiaxu112019}  needs  the  additional human  model to extract  the  human  landmark to help the model locate the key local feature.   So, the  structure  of PGFA~\cite{Jiaxu112019} is much  more complex  than  that of  our CBDB-Net. 
In  contrast,  our CBDB-Net  needn't any  auxiliary model,  and  is  a  kind  of simple and efficient person Re-ID model.

In addition to  the good performance in the two  kind person Re-ID   tasks,  we  believe that our CBDB-Net can be effective  in other image  retrieval  tasks.  We  thirdly  evaluate   the  performance   of our CBDB-Net  on the   clothes  retrieval task.  As  shown  in  Table~\ref{table4}, our CBDB-Net   also   achieves a good  performance  on the  In-shop clothes retrieval  dataset.

\begin{figure}\centering
	\begin{center} 
		\includegraphics[scale=0.4]{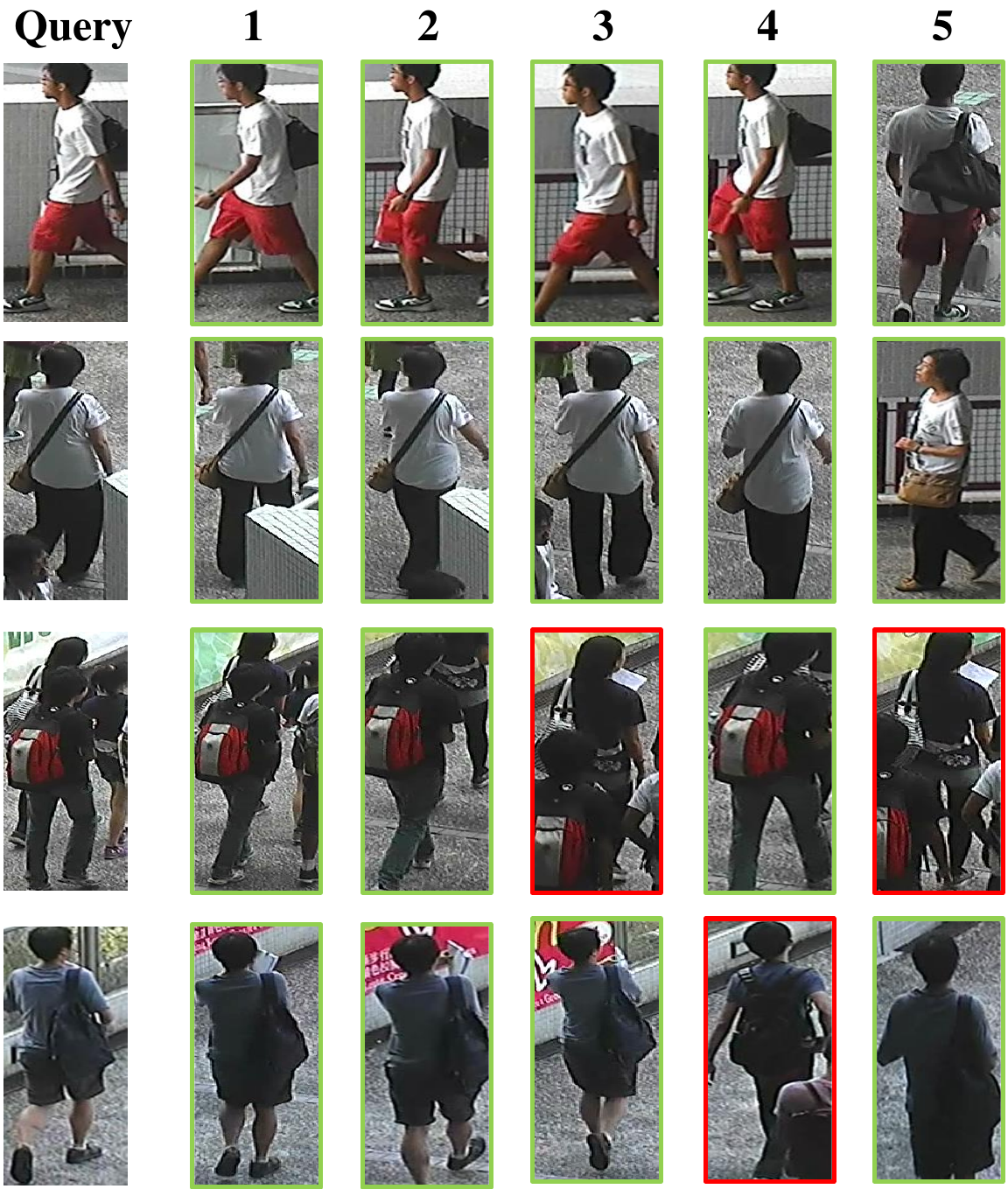}
	\end{center} 
	\caption{The top-5 ranking list for the query peson  on CUHK03-Detected text set from the proposed CBDB-Net. The correct  retrieval  results are highlighted by green borders and the incorrect retrieval  results by red borders. } 
	\label{Query-Person} 
\end{figure}

Some sample query results are illustrated in Figure~\ref{Query-Person}. We
can see that, given a  person image, CBDB-Net Network  can effectively  retrieve the same  person  images in the  gallery set.  However,  there are retrieval negative  results in Figure~\ref{Query-Person}. In  the third row of  Figure~\ref{Query-Person}, the  query person  was  also mixed in other person  images,  which made  the model confused  extract the suitable person feature.   In  the fourth row of  Figure~\ref{Query-Person}, the   occlusion  makes other features of the incorrectly retrieved perosn  body very similar to quary person image.  Thus,  it  is necessary  to further  improve  the CBDB-Net  in the occulsion person Re-ID task in future  works.  
Overall, our observations endorse the superiority of CBDB-Net by combing the ``Consecutive Batch DropBlock  Module''  and ``Elastic Loss''.
Compared  with other  state-of-the-art   approaches,  our  model  is simple  and effective  especially our  testing mode.  
In our CBDB-Net, we  only extract a  $512$ dimension  global descriptor to conduct  the person matching task  and gain good  performance.

\section{Conclusion}

In this paper, we propose a novel  person Re-ID model,  Consecutive  Batch DropBlock Network (CBDB-Net), to improve  the ability of the person Re-ID  model on capturing  the  robust  and high-quality feature descriptor  for  the person  matching.  
Specifically,  firstly   Consecutive  Batch DropBlock  Module  is  proposed  to  exploit  multiple  incomplete  descriptors, which can  effectively    push the person Re-ID model to  capture  the robust feature descriptor.  
Secondly,   the  Elastic Loss  is   designed to adaptively  mine and balance  the hard  sample pairs in the  training process.  
Extensive experiments show that our CBDB-Net achieves the competitive  performance on  three  generic  person Re-ID  datasets,  three  occlusion person  Re-ID datasets, and the generic  image retrieval  task.

\ifCLASSOPTIONcaptionsoff
\newpage
\fi

{\small
	\bibliographystyle{ieee_fullname}
	\bibliography{mybibfile}
}



\end{document}